%% file: main.tex
\documentclass[runningheads]{llncs}
\usepackage{graphicx}
\usepackage{tikz}
\usepackage{ulem}%ye
\usepackage{tabu}
\usepackage{adjustbox}
\usepackage{booktabs}

\usepackage{booktabs}
\usepackage{adjustbox}
\usepackage{array}
\usepackage{placeins}
\usepackage{afterpage}
\usepackage{stfloats}
\usepackage{textcomp}

\usepackage{comment}
\usepackage{amsmath,amssymb} % define this before the line numbering.
\usepackage{color}
\usepackage{multicol}
\usepackage{multirow}
\usepackage{floatrow}
\usepackage{subcaption,booktabs}
\usepackage[misc]{ifsym}
\usepackage[width=122mm,left=12mm,paperwidth=146mm,height=193mm,top=12mm,paperheight=217mm]{geometry}

\newcommand{\ourMethod}{GitNet}%yexiaoqing 
\usepackage[font=normalsize,labelfont=bf]{caption}
\begin{document}
% \linenumbers
\pagestyle{headings}
\mainmatter
\def\ECCVSubNumber{2449}  % Insert your submission number here

\newcommand{\tanxiao}[1]{{\color{black} #1}}
\newcommand{\Xiaoqing}[1]{{\color{black}{#1}}}
\newcommand{\old}[1]{{\color{gray} \sout{#1}}} %
\newcommand{\printfnsymbol}[1]{%
  \textsuperscript{\@fnsymbol{#1}}%
}
\title{GitNet: Geometric Prior-based Transformation for Birds-Eye-View Segmentation } % Replace with your title

%\begin{comment}
% \titlerunning{ECCV-22 submission ID \ECCVSubNumber} 
% \authorrunning{ECCV-22 submission ID \ECCVSubNumber} 
% \author{Anonymous ECCV submission}
% \institute{Paper ID \ECCVSubNumber}
%\end{comment}
%******************
\include{math_commands}
% \begin{comment}
\titlerunning{Geometric Prior-based Transformation for BEV Segmentation}
% \author{Shi Gong\thanks{Authors contribute equally.}\inst{1} \and
% Xiaoqing Ye\inst{\star 2}\orcidID{0000−0003−3268−880X} \and
% Xiao Tan\inst{2}\orcidID{0000-0002-4888-4445} \and
% Jingdong Wang\inst{2}\orcidID{0000-0002-4888-4445} \and
% Errui Ding\inst{2}\orcidID{0000-0002-1867-5378} \and
% Yu Zhou\thanks{Corresponding author.}\inst{1} \and
% Xiang Bai\inst{1}\orcidID{0000−0002−3449−5940}
% }
\author{Shi Gong\thanks{Authors contribute equally.}\inst{1} \and
Xiaoqing Ye\inst{\star 2} \and
Xiao Tan\inst{2}\and
Jingdong Wang\inst{2} \and \\
Errui Ding\inst{2} \and
Yu Zhou\thanks{Corresponding author.}\inst{1} \and
Xiang Bai\inst{1}
}

\authorrunning{Gong et al.}
% \institute{Huazhong University of Science and Technology \and
% \end{comment}

\institute{Huazhong University of Science and Technology \\ \{gongshi,yuzhou\}@hust.edu.cn \and Baidu Inc., China
}

%******************
\maketitle

\begin{abstract}
Birds-eye-view (BEV) semantic segmentation is critical for autonomous driving for its powerful spatial representation ability. It is challenging to estimate the BEV semantic maps from monocular images due to the spatial gap, since it is implicitly required to realize both the perspective-to-BEV transformation and segmentation.
We present a novel two-stage \textbf{G}eometry Pr\textbf{I}or-based \textbf{T}ransformation framework named GitNet, consisting of (i) the geometry-guided pre-alignment and (ii) ray-based transformer. 
In the first stage, we decouple the BEV segmentation into the perspective image segmentation and geometric prior-based mapping, with explicit supervision by projecting the BEV semantic labels onto the image plane to learn visibility-aware features and learnable geometry to translate into BEV space. Second, the pre-aligned coarse BEV features are further deformed by ray-based transformers to take visibility knowledge into account.
GitNet achieves the leading performance on the challenging nuScenes and Argoverse Datasets. 

\keywords{Birds-Eye-View, Segmentation, Geometric Prior-based}
\end{abstract}

\input{sections/01_Introduction}

\input{sections/02_Relatedwork}

\input{sections/03_Method}

\input{sections/04_Experiments}

\input{sections/05_Conclusion}
\clearpage

\bibliographystyle{splncs04}
\bibliography{egbib}

\clearpage
\input{sections/06_Appendix}

\end{document}

%% file: math_commands.tex
\newcommand{\denc}{d_{\rm enc}}
\newcommand{\ddec}{d_{\rm dec}}
\renewcommand{\Re}{\mathbb{R}}
\newcommand{\hy}{\hat{y}}
\newcommand{\hb}{\hat{b}}
\newcommand{\hp}{\hat{p}}
\newcommand{\ty}{\tilde{y}}

\renewcommand{\Sigma}{\mathfrak{S}}

\newcommand{\mhattn}{\text{\rm multi-head-attn}}
\newcommand{\mhsattn}{\text{\rm multi-head-self-attn}}
\newcommand{\attn}{\text{\rm attn}}
\newcommand{\xq}{X_{\rm q}} %
\newcommand{\xqout}{\tilde{X}_{\rm q}} %
\newcommand{\xqbb}{X'_{\rm q}} %

\newcommand{\xkv}{X_{\rm kv}} %
\newcommand{\weit}{T} %
\newcommand{\Nq}{N_{\rm q}} %
\newcommand{\Nkv}{N_{\rm kv}} %
\newcommand{\ques}{Q}
\newcommand{\keys}{K}
\newcommand{\vals}{V}
\newcommand{\posq}{P_{\rm q}} %
\newcommand{\poskv}{P_{\rm kv}} %
\newcommand{\proj}{L}

\newcommand{\dmodel}{d}
\newcommand{\dk}{d'}

\newcommand{\indic}[1]{\mathds{1}_{\{#1\}}}

\newcommand{\loss}[1]{{\cal L}(#1)}
\newcommand{\closs}[1]{{\cal L}_{\rm class}(#1)}
\newcommand{\bloss}[1]{{\cal L}_{\rm box}(#1)}
\newcommand{\iouloss}[1]{{\cal L}_{\rm iou}(#1)}
\newcommand{\diceloss}[1]{{\cal L}_{\rm DICE}(#1)}
\newcommand{\hloss}[1]{{\cal L}_{\rm Hungarian}(#1)}

\newcommand{\lmatch}[1]{{\cal L}_{\rm match}(#1)}

\newcommand{\figleft}{{\em (Left)}}
\newcommand{\figcenter}{{\em (Center)}}
\newcommand{\figright}{{\em (Right)}}
\newcommand{\figtop}{{\em (Top)}}
\newcommand{\figbottom}{{\em (Bottom)}}
\newcommand{\captiona}{{\em (a)}}
\newcommand{\captionb}{{\em (b)}}
\newcommand{\captionc}{{\em (c)}}
\newcommand{\captiond}{{\em (d)}}

\newcommand{\newterm}[1]{{\bf #1}}

\def\figref#1{figure~\ref{#1}}
\def\Figref#1{Figure~\ref{#1}}
\def\twofigref#1#2{figures \ref{#1} and \ref{#2}}
\def\quadfigref#1#2#3#4{figures \ref{#1}, \ref{#2}, \ref{#3} and \ref{#4}}
\def\secref#1{section~\ref{#1}}
\def\Secref#1{Section~\ref{#1}}
\def\twosecrefs#1#2{sections \ref{#1} and \ref{#2}}
\def\secrefs#1#2#3{sections \ref{#1}, \ref{#2} and \ref{#3}}
\def\eqref#1{equation~\ref{#1}}
\def\Eqref#1{Equation~\ref{#1}}
\def\plaineqref#1{\ref{#1}}
\def\chapref#1{chapter~\ref{#1}}
\def\Chapref#1{Chapter~\ref{#1}}
\def\rangechapref#1#2{chapters\ref{#1}--\ref{#2}}
\def\algref#1{algorithm~\ref{#1}}
\def\Algref#1{Algorithm~\ref{#1}}
\def\twoalgref#1#2{algorithms \ref{#1} and \ref{#2}}
\def\Twoalgref#1#2{Algorithms \ref{#1} and \ref{#2}}
\def\partref#1{part~\ref{#1}}
\def\Partref#1{Part~\ref{#1}}
\def\twopartref#1#2{parts \ref{#1} and \ref{#2}}

\def\ceil#1{\lceil #1 \rceil}
\def\floor#1{\lfloor #1 \rfloor}
\def\1{\bm{1}}
\newcommand{\train}{\mathcal{D}}
\newcommand{\valid}{\mathcal{D_{\mathrm{valid}}}}
\newcommand{\test}{\mathcal{D_{\mathrm{test}}}}

\def\eps{{\epsilon}}

\def\reta{{\textnormal{$\eta$}}}
\def\ra{{\textnormal{a}}}
\def\rb{{\textnormal{b}}}
\def\rc{{\textnormal{c}}}
\def\rd{{\textnormal{d}}}
\def\re{{\textnormal{e}}}
\def\rf{{\textnormal{f}}}
\def\rg{{\textnormal{g}}}
\def\rh{{\textnormal{h}}}
\def\ri{{\textnormal{i}}}
\def\rj{{\textnormal{j}}}
\def\rk{{\textnormal{k}}}
\def\rl{{\textnormal{l}}}
\def\rn{{\textnormal{n}}}
\def\ro{{\textnormal{o}}}
\def\rp{{\textnormal{p}}}
\def\rq{{\textnormal{q}}}
\def\rr{{\textnormal{r}}}
\def\rs{{\textnormal{s}}}
\def\rt{{\textnormal{t}}}
\def\ru{{\textnormal{u}}}
\def\rv{{\textnormal{v}}}
\def\rw{{\textnormal{w}}}
\def\rx{{\textnormal{x}}}
\def\ry{{\textnormal{y}}}
\def\rz{{\textnormal{z}}}

\def\rvepsilon{{\mathbf{\epsilon}}}
\def\rvtheta{{\mathbf{\theta}}}
\def\rva{{\mathbf{a}}}
\def\rvb{{\mathbf{b}}}
\def\rvc{{\mathbf{c}}}
\def\rvd{{\mathbf{d}}}
\def\rve{{\mathbf{e}}}
\def\rvf{{\mathbf{f}}}
\def\rvg{{\mathbf{g}}}
\def\rvh{{\mathbf{h}}}
\def\rvu{{\mathbf{i}}}
\def\rvj{{\mathbf{j}}}
\def\rvk{{\mathbf{k}}}
\def\rvl{{\mathbf{l}}}
\def\rvm{{\mathbf{m}}}
\def\rvn{{\mathbf{n}}}
\def\rvo{{\mathbf{o}}}
\def\rvp{{\mathbf{p}}}
\def\rvq{{\mathbf{q}}}
\def\rvr{{\mathbf{r}}}
\def\rvs{{\mathbf{s}}}
\def\rvt{{\mathbf{t}}}
\def\rvu{{\mathbf{u}}}
\def\rvv{{\mathbf{v}}}
\def\rvw{{\mathbf{w}}}
\def\rvx{{\mathbf{x}}}
\def\rvy{{\mathbf{y}}}
\def\rvz{{\mathbf{z}}}

\def\erva{{\textnormal{a}}}
\def\ervb{{\textnormal{b}}}
\def\ervc{{\textnormal{c}}}
\def\ervd{{\textnormal{d}}}
\def\erve{{\textnormal{e}}}
\def\ervf{{\textnormal{f}}}
\def\ervg{{\textnormal{g}}}
\def\ervh{{\textnormal{h}}}
\def\ervi{{\textnormal{i}}}
\def\ervj{{\textnormal{j}}}
\def\ervk{{\textnormal{k}}}
\def\ervl{{\textnormal{l}}}
\def\ervm{{\textnormal{m}}}
\def\ervn{{\textnormal{n}}}
\def\ervo{{\textnormal{o}}}
\def\ervp{{\textnormal{p}}}
\def\ervq{{\textnormal{q}}}
\def\ervr{{\textnormal{r}}}
\def\ervs{{\textnormal{s}}}
\def\ervt{{\textnormal{t}}}
\def\ervu{{\textnormal{u}}}
\def\ervv{{\textnormal{v}}}
\def\ervw{{\textnormal{w}}}
\def\ervx{{\textnormal{x}}}
\def\ervy{{\textnormal{y}}}
\def\ervz{{\textnormal{z}}}

\def\rmA{{\mathbf{A}}}
\def\rmB{{\mathbf{B}}}
\def\rmC{{\mathbf{C}}}
\def\rmD{{\mathbf{D}}}
\def\rmE{{\mathbf{E}}}
\def\rmF{{\mathbf{F}}}
\def\rmG{{\mathbf{G}}}
\def\rmH{{\mathbf{H}}}
\def\rmI{{\mathbf{I}}}
\def\rmJ{{\mathbf{J}}}
\def\rmK{{\mathbf{K}}}
\def\rmL{{\mathbf{L}}}
\def\rmM{{\mathbf{M}}}
\def\rmN{{\mathbf{N}}}
\def\rmO{{\mathbf{O}}}
\def\rmP{{\mathbf{P}}}
\def\rmQ{{\mathbf{Q}}}
\def\rmR{{\mathbf{R}}}
\def\rmS{{\mathbf{S}}}
\def\rmT{{\mathbf{T}}}
\def\rmU{{\mathbf{U}}}
\def\rmV{{\mathbf{V}}}
\def\rmW{{\mathbf{W}}}
\def\rmX{{\mathbf{X}}}
\def\rmY{{\mathbf{Y}}}
\def\rmZ{{\mathbf{Z}}}

\def\ermA{{\textnormal{A}}}
\def\ermB{{\textnormal{B}}}
\def\ermC{{\textnormal{C}}}
\def\ermD{{\textnormal{D}}}
\def\ermE{{\textnormal{E}}}
\def\ermF{{\textnormal{F}}}
\def\ermG{{\textnormal{G}}}
\def\ermH{{\textnormal{H}}}
\def\ermI{{\textnormal{I}}}
\def\ermJ{{\textnormal{J}}}
\def\ermK{{\textnormal{K}}}
\def\ermL{{\textnormal{L}}}
\def\ermM{{\textnormal{M}}}
\def\ermN{{\textnormal{N}}}
\def\ermO{{\textnormal{O}}}
\def\ermP{{\textnormal{P}}}
\def\ermQ{{\textnormal{Q}}}
\def\ermR{{\textnormal{R}}}
\def\ermS{{\textnormal{S}}}
\def\ermT{{\textnormal{T}}}
\def\ermU{{\textnormal{U}}}
\def\ermV{{\textnormal{V}}}
\def\ermW{{\textnormal{W}}}
\def\ermX{{\textnormal{X}}}
\def\ermY{{\textnormal{Y}}}
\def\ermZ{{\textnormal{Z}}}

\def\vzero{{\bm{0}}}
\def\vone{{\bm{1}}}
\def\vmu{{\bm{\mu}}}
\def\vtheta{{\bm{\theta}}}
\def\va{{\bm{a}}}
\def\vb{{\bm{b}}}
\def\vc{{\bm{c}}}
\def\vd{{\bm{d}}}
\def\ve{{\bm{e}}}
\def\vf{{\bm{f}}}
\def\vg{{\bm{g}}}
\def\vh{{\bm{h}}}
\def\vi{{\bm{i}}}
\def\vj{{\bm{j}}}
\def\vk{{\bm{k}}}
\def\vl{{\bm{l}}}
\def\vm{{\bm{m}}}
\def\vn{{\bm{n}}}
\def\vo{{\bm{o}}}
\def\vp{{\bm{p}}}
\def\vq{{\bm{q}}}
\def\vr{{\bm{r}}}
\def\vs{{\bm{s}}}
\def\vt{{\bm{t}}}
\def\vu{{\bm{u}}}
\def\vv{{\bm{v}}}
\def\vw{{\bm{w}}}
\def\vx{{\bm{x}}}
\def\vy{{\bm{y}}}
\def\vz{{\bm{z}}}

\def\evalpha{{\alpha}}
\def\evbeta{{\beta}}
\def\evepsilon{{\epsilon}}
\def\evlambda{{\lambda}}
\def\evomega{{\omega}}
\def\evmu{{\mu}}
\def\evpsi{{\psi}}
\def\evsigma{{\sigma}}
\def\evtheta{{\theta}}
\def\eva{{a}}
\def\evb{{b}}
\def\evc{{c}}
\def\evd{{d}}
\def\eve{{e}}
\def\evf{{f}}
\def\evg{{g}}
\def\evh{{h}}
\def\evi{{i}}
\def\evj{{j}}
\def\evk{{k}}
\def\evl{{l}}
\def\evm{{m}}
\def\evn{{n}}
\def\evo{{o}}
\def\evp{{p}}
\def\evq{{q}}
\def\evr{{r}}
\def\evs{{s}}
\def\evt{{t}}
\def\evu{{u}}
\def\evv{{v}}
\def\evw{{w}}
\def\evx{{x}}
\def\evy{{y}}
\def\evz{{z}}

\def\mA{{\bm{A}}}
\def\mB{{\bm{B}}}
\def\mC{{\bm{C}}}
\def\mD{{\bm{D}}}
\def\mE{{\bm{E}}}
\def\mF{{\bm{F}}}
\def\mG{{\bm{G}}}
\def\mH{{\bm{H}}}
\def\mI{{\bm{I}}}
\def\mJ{{\bm{J}}}
\def\mK{{\bm{K}}}
\def\mL{{\bm{L}}}
\def\mM{{\bm{M}}}
\def\mN{{\bm{N}}}
\def\mO{{\bm{O}}}
\def\mP{{\bm{P}}}
\def\mQ{{\bm{Q}}}
\def\mR{{\bm{R}}}
\def\mS{{\bm{S}}}
\def\mT{{\bm{T}}}
\def\mU{{\bm{U}}}
\def\mV{{\bm{V}}}
\def\mW{{\bm{W}}}
\def\mX{{\bm{X}}}
\def\mY{{\bm{Y}}}
\def\mZ{{\bm{Z}}}
\def\mBeta{{\bm{\beta}}}
\def\mPhi{{\bm{\Phi}}}
\def\mLambda{{\bm{\Lambda}}}
\def\mSigma{{\bm{\Sigma}}}

% \DeclareMathAlphabet{\mathsfit}{\encodingdefault}{\sfdefault}{m}{sl}
% \SetMathAlphabet{\mathsfit}{bold}{\encodingdefault}{\sfdefault}{bx}{n}
\newcommand{\tens}[1]{\bm{\mathsfit{#1}}}
\def\tA{{\tens{A}}}
\def\tB{{\tens{B}}}
\def\tC{{\tens{C}}}
\def\tD{{\tens{D}}}
\def\tE{{\tens{E}}}
\def\tF{{\tens{F}}}
\def\tG{{\tens{G}}}
\def\tH{{\tens{H}}}
\def\tI{{\tens{I}}}
\def\tJ{{\tens{J}}}
\def\tK{{\tens{K}}}
\def\tL{{\tens{L}}}
\def\tM{{\tens{M}}}
\def\tN{{\tens{N}}}
\def\tO{{\tens{O}}}
\def\tP{{\tens{P}}}
\def\tQ{{\tens{Q}}}
\def\tR{{\tens{R}}}
\def\tS{{\tens{S}}}
\def\tT{{\tens{T}}}
\def\tU{{\tens{U}}}
\def\tV{{\tens{V}}}
\def\tW{{\tens{W}}}
\def\tX{{\tens{X}}}
\def\tY{{\tens{Y}}}
\def\tZ{{\tens{Z}}}

\def\gA{{\mathcal{A}}}
\def\gB{{\mathcal{B}}}
\def\gC{{\mathcal{C}}}
\def\gD{{\mathcal{D}}}
\def\gE{{\mathcal{E}}}
\def\gF{{\mathcal{F}}}
\def\gG{{\mathcal{G}}}
\def\gH{{\mathcal{H}}}
\def\gI{{\mathcal{I}}}
\def\gJ{{\mathcal{J}}}
\def\gK{{\mathcal{K}}}
\def\gL{{\mathcal{L}}}
\def\gM{{\mathcal{M}}}
\def\gN{{\mathcal{N}}}
\def\gO{{\mathcal{O}}}
\def\gP{{\mathcal{P}}}
\def\gQ{{\mathcal{Q}}}
\def\gR{{\mathcal{R}}}
\def\gS{{\mathcal{S}}}
\def\gT{{\mathcal{T}}}
\def\gU{{\mathcal{U}}}
\def\gV{{\mathcal{V}}}
\def\gW{{\mathcal{W}}}
\def\gX{{\mathcal{X}}}
\def\gY{{\mathcal{Y}}}
\def\gZ{{\mathcal{Z}}}

\def\sA{{\mathbb{A}}}
\def\sB{{\mathbb{B}}}
\def\sC{{\mathbb{C}}}
\def\sD{{\mathbb{D}}}
\def\sF{{\mathbb{F}}}
\def\sG{{\mathbb{G}}}
\def\sH{{\mathbb{H}}}
\def\sI{{\mathbb{I}}}
\def\sJ{{\mathbb{J}}}
\def\sK{{\mathbb{K}}}
\def\sL{{\mathbb{L}}}
\def\sM{{\mathbb{M}}}
\def\sN{{\mathbb{N}}}
\def\sO{{\mathbb{O}}}
\def\sP{{\mathbb{P}}}
\def\sQ{{\mathbb{Q}}}
\def\sR{{\mathbb{R}}}
\def\sS{{\mathbb{S}}}
\def\sT{{\mathbb{T}}}
\def\sU{{\mathbb{U}}}
\def\sV{{\mathbb{V}}}
\def\sW{{\mathbb{W}}}
\def\sX{{\mathbb{X}}}
\def\sY{{\mathbb{Y}}}
\def\sZ{{\mathbb{Z}}}

\def\emLambda{{\Lambda}}
\def\emA{{A}}
\def\emB{{B}}
\def\emC{{C}}
\def\emD{{D}}
\def\emE{{E}}
\def\emF{{F}}
\def\emG{{G}}
\def\emH{{H}}
\def\emI{{I}}
\def\emJ{{J}}
\def\emK{{K}}
\def\emL{{L}}
\def\emM{{M}}
\def\emN{{N}}
\def\emO{{O}}
\def\emP{{P}}
\def\emQ{{Q}}
\def\emR{{R}}
\def\emS{{S}}
\def\emT{{T}}
\def\emU{{U}}
\def\emV{{V}}
\def\emW{{W}}
\def\emX{{X}}
\def\emY{{Y}}
\def\emZ{{Z}}
\def\emSigma{{\Sigma}}

\newcommand{\etens}[1]{\mathsfit{#1}}
\def\etLambda{{\etens{\Lambda}}}
\def\etA{{\etens{A}}}
\def\etB{{\etens{B}}}
\def\etC{{\etens{C}}}
\def\etD{{\etens{D}}}
\def\etE{{\etens{E}}}
\def\etF{{\etens{F}}}
\def\etG{{\etens{G}}}
\def\etH{{\etens{H}}}
\def\etI{{\etens{I}}}
\def\etJ{{\etens{J}}}
\def\etK{{\etens{K}}}
\def\etL{{\etens{L}}}
\def\etM{{\etens{M}}}
\def\etN{{\etens{N}}}
\def\etO{{\etens{O}}}
\def\etP{{\etens{P}}}
\def\etQ{{\etens{Q}}}
\def\etR{{\etens{R}}}
\def\etS{{\etens{S}}}
\def\etT{{\etens{T}}}
\def\etU{{\etens{U}}}
\def\etV{{\etens{V}}}
\def\etW{{\etens{W}}}
\def\etX{{\etens{X}}}
\def\etY{{\etens{Y}}}
\def\etZ{{\etens{Z}}}

\newcommand{\pdata}{p_{\rm{data}}}
\newcommand{\ptrain}{\hat{p}_{\rm{data}}}
\newcommand{\Ptrain}{\hat{P}_{\rm{data}}}
\newcommand{\pmodel}{p_{\rm{model}}}
\newcommand{\Pmodel}{P_{\rm{model}}}
\newcommand{\ptildemodel}{\tilde{p}_{\rm{model}}}
\newcommand{\pencode}{p_{\rm{encoder}}}
\newcommand{\pdecode}{p_{\rm{decoder}}}
\newcommand{\precons}{p_{\rm{reconstruct}}}

\newcommand{\laplace}{\mathrm{Laplace}} %

\newcommand{\E}{\mathbb{E}}
\newcommand{\Ls}{\mathcal{L}}
\newcommand{\R}{\mathbb{R}}
\newcommand{\emp}{\tilde{p}}
\newcommand{\lr}{\alpha}
\newcommand{\reg}{\lambda}
\newcommand{\rect}{\mathrm{rectifier}}
\newcommand{\softmax}{\mathrm{softmax}}
\newcommand{\sigmoid}{\sigma}
\newcommand{\softplus}{\zeta}
\newcommand{\KL}{D_{\mathrm{KL}}}
\newcommand{\Var}{\mathrm{Var}}
\newcommand{\standarderror}{\mathrm{SE}}
\newcommand{\Cov}{\mathrm{Cov}}
\newcommand{\normlzero}{L^0}
\newcommand{\normlone}{L^1}
\newcommand{\normltwo}{L^2}
\newcommand{\normlp}{L^p}
\newcommand{\normmax}{L^\infty}

\newcommand{\parents}{Pa} %

\let\ab\allowbreak

%% file: sections/01_Introduction.tex
\section{Introduction}
\label{sec:introduction}
The birds-eye-view (BEV) semantic map is a compact representation of the surrounding environment for autonomous driving, which provides both the layout of road elements and the occupancy of objects. Such representations are useful for downstream tasks such as path planning, collision avoidance. In this work, we focus on BEV map estimation from monocular images.

The BEV semantic segmentation is particularly challenging for two reasons. 
First, the BEV segmentation implicitly involves two coupled tasks: mapping from perspective view to the birds-eye-view, and pixel-wise classification. \tanxiao{
Most existing methods \cite{roddick20_PON,lu2019_VED,pan2020_VPN,philion2020_lift,mani2020_monolayout,saha21_sta} learn to convert the image features from the perspective view to the BEV and then perform segmentation. The training process is supervised by the loss function defined in the BEV space alone, and thus the learning procedure of mapping and pixel-wise classification is coupled in these approaches. How to explicitly incorporate the geometry prior knowledge to decouple the feature for mapping and classification }\Xiaoqing{remains unexplored.} 
\tanxiao{Secondly, a fundamental difference between monocular image segmentation and BEV segmentation lies in that the latter requires inferring the labels of occluded objects behind foreground objects, which places a tremendous difficulty for the network to learn effective feature representation to differentiate the invisible from the visible.} In the previous IPM-based methods \cite{sengupta2012_Auto,zhu2018_generative,reiher2020_sim2real}, the features of foreground visible objects occupy the invisible regions in the BEV space. Since the visibility of pixels 
is not encoded in the features, it is tough for a convolutional neural network to recover the missing information in the invisible regions.
%  And the recovering of invisible region needs rich contextual information, the convolutional network 
% and mapping from perspective view to estimating the depth of pixels. Since inferring depth from a monocular image is an ill-posed problem, some priors, such as flatten ground, camera height and intrinsics, are essential to recover the depth. 
% 
 %the  dealing with the occlusion is difficult for the BEV segmentation, such as road being partially occluded by the foreground objects, or self-occlusions of objects. Effective BEV segmentation methods need learn to reason about the classes of occluded parts.
% 
% On the other hand, the inverse perspective mapping (IPM) based methods \cite{sengupta2012_Auto,reiher2020_sim2real,zhu2018_generative}, leverage the prior knowledge to lift the pixels to the BEV, but they fail to handle the occlusion.

%On the one hand, most learning-based methods \cite{roddick20_PON,lu2019_VED,pan2020_VPN} directly learn a mapping from image space to BEV space without explicitly incorporating the prior knowledge into the model. It is suboptimal for the convergence of model training. On the other hand, the inverse perspective mapping (IPM) based methods \cite{sengupta2012_Auto,reiher2020_sim2real,zhu2018_generative}, leverage the prior knowledge to lift the pixels to the BEV, but they fail to handle the occlusion.
% 
To address the aforementioned concerns, we derive a novel two-stage transformation from the perspective space to the BEV space.
In the first stage, we leverage the proposed Geometry-guided Pre-Alignment (GPA) to obtain coarse pre-aligned BEV features.
In the GPA, we decouple the BEV segmentation into the perspective image segmentation and geometric prior-based mapping, with explicit supervision by projecting the BEV semantic labels onto the image plane. As the projected labels reflect all ground regions including visible and invisible ones in the perspective view, while the perspective image appearance features only reflect the visible regions, we obtain the visibility-aware image features by fusing the information of projected labels and appearance features. We warp the visibility-aware features into BEV space via the learnable geometry.

In the second stage, the pre-aligned BEV features are further enhanced by the proposed Ray-based Transformer (RT), which adopts the efficient ray-based attention mechanism that we compute the attention map in a single column so as to keep the high-resolution of feature maps.
The pre-aligned BEV features conveying \textit{appearance} and \textit{visibility} information, along with BEV positional encoding, work as \textit{Queries}, and the augmented perspective features serve as \textit{Keys} and \textit{Values}. 
%Different from most existing methods that compute full-image attention, the Ray-based Transformer performs attention within the same column, so that we can perform on high-resolution feature maps, which is indispensable in pixel-wise classification. 
% 
Cooperating with the projected labels, the novel Depth-Aware Dice loss is proposed to alleviate the dominant effect by closer instances in perspective view. Besides, since
% the layouts are easier to be correctly segmented and  most pixels
those pixels that have easily-classified appearances or follow a simple perspective-to-BEV mapping, such as most road regions, comprise the majority of the loss,
% since the layouts are easier to be correctly segmented since they are on the ground plane whereas objects above the ground are hard to be distinguished, 
we present a Self-Weighted Dice loss to balance the easy-hard samples among categories.
% visibility-aware and pre-aligned, and they are deformed by aggregating appearance information, which is instantiated by our transformer-based module. 
% incorporate the geometric prior to \textit{Geometry-guided Pre-Alignment} module.
% which converts the BEV semantic labels to the perspective image plane, the perspective features to the BEV space by inverse projection with the learned camera height. For better alignment between the perspective space and the birds-eye-view space, the projected BEV labels are used to modulate the perspective features to dig out the underlying features of occluded objects. 
% 
% After obtaining the pre-aligned BEV features, we further adopt the \textit{Ray-based Transformer} module as the second stage, 
% 
% In the \textit{Ray-based Transformer}, 
% 
To sum up, the main contributions of our work are as follows:
\vspace{-\topsep}
\begin{itemize}
\setlength{\parskip}{0pt}
\setlength{\itemsep}{0pt plus 1pt}
% \begin{itemize}
% \vspace{2mm}
\item We propose a novel two-stage transformation from perspective view to birds-eye-view. In the first stage, we decouple the BEV segmentation into the perspective image segmentation and geometric prior-based mapping, and provide visibility-aware and pre-aligned BEV features. 
In the second stage, the warped features are deformed by aggregating appearance information.
\item We introduce a Depth-aware Dice loss that removes the perspective effect on the perspective image segmentation and a Self-weighted Dice loss to re-weight the easy-hard samples.
\item Our framework presents new state-of-the-art performance on two large-scale datasets including nuScenes and Argoverse.
\end{itemize}

%% file: sections/02_Relatedwork.tex
\section{Related work}
% Relatively few works however have tackled the more specific problem of generating semantic maps from images.
% Many prior works have trackled the inherently ill-posed problem of lifting 2D perspective images into a birds-eye-view representation. \cite{reiher2020_sim2real} dealt specifically with the problem of generating semantic BEV maps directly from images and used simulator to obtain the ground truth.
% Recent multi-sensor datasets, such as NuScenes\cite{caesar2020_nus} or Argoverse \cite{chang2019_argo}, made it possible to directly supervise models on real-world data by generating bird’s-eye view semantic segmentation labels from 3D object detections.
% In this section, we survey the related literature on birds-eye-view semantic segmentation. The key point of inferring semantic BEV maps is tranforming the representation from the image to birds-eye-view. According to the paradigm of transforming, the prior methods can be grouped into four categories: (\romannumeral1) Inverse Perspective Mapping (IPM) based methods; (\romannumeral2) Depth based methods; (\romannumeral3) Bottleneck based methods; (\romannumeral4) Attention based method. 
% Recently, the BEV semantic segmentation has attracted much attention because it provides a compact representation that captures the spatial configuration of road scenes. 
Most BEV segmentation works follow the similar pipeline to first extract  features from the monocular image, and then convert the features from the perspective view (PV) to the birds-eye-view (BEV). Based on different PV-BEV transformation strategies, the methods can be grouped into four categories as follows:
% : (\romannumeral1) inverse perspective mapping (IPM) based methods; (\romannumeral2) depth based methods; (\romannumeral3) bottleneck based methods; (\romannumeral4) attention based method.

\noindent\textbf{IPM-based Methods:} 
% \cite{sengupta2012_Auto} first introduce the problem of visual semantic mapping, using the IPM to lift the image segmentation to the BEV ground plane.  OFT\cite{roddick2018_OFT}
An early work \cite{sengupta2012_Auto} performs semantic segmentation in the image plane and then transforms the semantic results into the BEV space via a homography. This approach works well for predicting flat road layout but fails for objects such as cars that stand above the ground plane. \cite{zhu2018_generative} alleviates this problem by training a generative adversarial network to refine the predictions from the IPM. More recently  \cite{reiher2020_sim2real} transforms the image features into BEV, which is then fed into a deep segmentation network for further refinement.
%In our work, a similar IPM technique is used to obtain coarse BEV features with the jitter of camera height taken into consideration, and the image pyramid features are column-wisely transformed to the BEV of different depth intervals. 
% and images features are explicitly supervised by the proposed depth-aware Dice loss.

\noindent\textbf{Depth-based methods:} 
The depth-based methods are one of the main streams in this field. \cite{henriques2018mapnet} adopts RGB-D images to learn an implicit representation for 3D localization. \cite{schulter2018_learning} leverages an in-painting CNN to infer the semantic labels and depths of the scene to obtain the BEV map by projecting the produced semantic point cloud onto the ground plane. EPOSH \cite{dwivedi2021_EPOSH} first performs monocular depth estimation and then exploits depth maps to transform 2D image features to the BEV space. \cite{philion2020_lift,reading2021_categorical,hu2021_fiery,saha21_sta} learn a depth distribution within pixels to lift 2D images to 3D point clouds, and then project the point clouds onto BEV space. 
% The depth based methods learn the intermediate representation of 3D voxel by 

\noindent\textbf{Bottleneck-based methods:}
% VPN\cite{pan2020_VPN} VED\cite{lu2019_VED} PON\cite{roddick20_PON}
% monolayout\cite{mani2020_monolayout}
VED \cite{lu2019_VED} uses the fully-connected bottleneck to realize the transformation, which loses the spatial information. Therefore the output is fairly coarse and fails to segment small objects. VPN \cite{pan2020_VPN} predicts the semantic BEV map from a stack of surround-view images, via a fully-connected view-transformer module.
 PON \cite{roddick20_PON} proposes a column-wise fully-connected layer to realize the transformation of features from image space to BEV space.  

\noindent\textbf{Attention based methods:}
The attention-based methods are attracting increasing attention. NEAT~\cite{chitta2021_Neat} proposes a novel representation termed neural attention fields, which compresses 2D image features into the BEV representation based on the attention map. TIM~\cite{Saha21_tim} transforms image columns to BEV polar rays via cross-attention. Though similar to our work, the BEV features are initialized to constant, and the geometric prior is not exploited in their method, which limits the capacity of reasoning in 3D space.

%% file: sections/03_Method.tex
\section{Method}
\label{text:method}
%ye
In this section, we first briefly present our \ourMethod{} approach, which learns the birds-eye-view (BEV) segmentation map from a  monocular image $I \in \mathbb{R}^{H\times W\times 3}$. The predicted BEV semantic map $S\in \mathbb{R}^{ Z \times X \times C}$ is in the ego camera coordinate, with  $Z$ and $X$ are the spatial dimensions of the regular lattice grid in BEV space and $C$ is the number of semantic categories including road layout and objects.

\subsection{Overview}
\label{text:overview}
\begin{figure}[t]
	\begin{center}
		\includegraphics[width=1\linewidth]{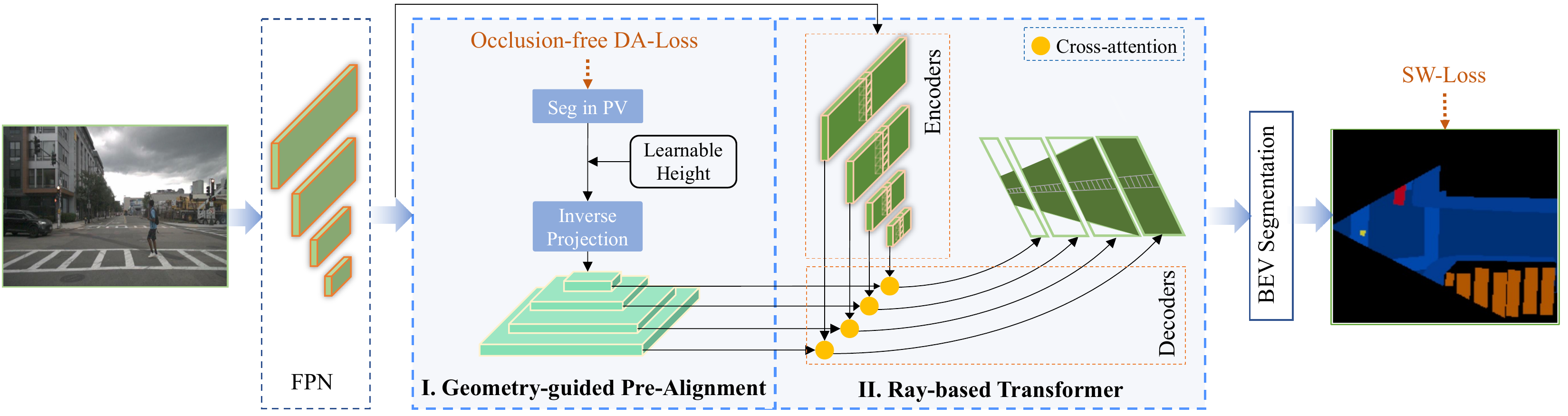}
	\end{center}
	%\vspace*{-5mm}
	\caption{The overview of the \ourMethod{} framework to predict the BEV semantic map from the perspective image.
 The multi-level pyramid image features extracted by the FPN are transformed to the BEV features by our \textit{two-stage} transformation pipeline, which includes the \textit{Geometry-guided Pre-Alignment} (\textit{GPA}) and \textit{Ray-based Transformer} (\textit{RT}). \Xiaoqing{The explicit supervision is enforced to the GPA Stage guided by the learnable camera height to learn visibility-aware features, which are then converted to pre-aligned BEV features.
%  and convert to pre-aligned BEV features
 }
%  \old{The \textit{GPA} supervises the segmentation on image by the occlusion-free DA-loss, and converts the image features to pre-aligned BEV features by the inverse projection with learnable camera height.} 
 The \textit{RT} column-wisely refines the PV features and pre-aligned BEV features with the mechanism of attention. The refined BEV features are fed into the BEV segmentation layers, which output $C$ pixel-wise binary classification.
	}
	%}
	\label{fig:overview}
\end{figure}

The goal of our network is to predict the semantic map of the scene on the birds-eye-view space from a monocular perspective image. The challenge of predicting the BEV semantic map lies in that the input and output representations exist in different spaces and thus the network is acquired to learn the transformation from perspective image view to orthographic BEV space.
As depicted in Fig. \ref{fig:overview}, our framework \Xiaoqing{is a two-stage pipeline that transforms the perspective view (PV) to the birds-eye-view.} It mainly consists of four modules, (i) the feature pyramid network (FPN) for multi-scale \Xiaoqing{perspective} feature representation, (ii) Geometry-guided Pre-Alignment that transfers features into BEV space based on the learnable camera height, (iii) the ray-based transformer module \Xiaoqing{for attention-based feature enhancement} before BEV segmentation, and (iv) the specially designed loss functions for re-weighting different pixels.

In our network, the core design is the two-stage transformation from the perspective space to the BEV space. Firstly we leverage the  {geometric guidance to provide \textit{appearance} and \textit{visibility} for initializing the transformed BEV features. To solve the ambiguity caused by the mounting height of the camera, we specially learn the height for better alignment between the perspective space and the birds-eye-view space. 
After obtaining the pre-aligned BEV features, we further adopt the ray-based transformer module \Xiaoqing{based on the column-wise attention for further enhancing the feature deformation in BEV space}
% \old{translating the coarse features into its corresponding BEV ray}  
for conducting semantic segmentation. 
% 
% \old{To alleviate the perspective effect caused by the imaging, we supervise the Geometric prior-based transformation module with Depth-Aware Dice (DA-Dice) loss }
\Xiaoqing{
% To explicitly enforce the GPA module to learn consistent perspective features with those in BEV space, we supervise the perspective features with projected labels from the BEV ground truth. 
The explicit supervision is enforced to the GPA Stage guided by the learnable camera height to learn visibility-aware features, which are then converted to pre-aligned BEV features.
In addition, to alleviate the perspective effect caused by the imaging, we organize the projected supervision loss in a depth-aware manner and further} propose the 
Self-Weighted Dice (SW-Dice) loss to re-weight the easy-hard samples. We will introduce the detailed design of each component in the following parts.}
% %%%%%%
% The FPN produces multi-scale image features, which are fed to downstream blocks for predicting the semantic BEV map of different depth intervals. 
% % 
% Segmentation is most commonly performed on a single-scale feature map. 
% However, in our case of BEV segmentation, because the perspective projection makes farther-away objects appear smaller, the construction of distant BEV maps needs high-resolution image features (low-level), and since close objects appear large in the image, we need large-receptive-field features maps (high-level).
% % 
% Formally we assign the depth interval ranging from $z_{k-1}$ to $z_k$, to the pyramid level $P_k$  by:
% \begin{equation}
%     z_k = \frac{f\Delta x}{s_k}
% \end{equation}
% % 
% where $s_k=2^{k+2}, k \in \{1,2,3,4\}$ is the downsampling factors of feature pyramid, $f$ the focal of camera and $\Delta x$ is the BEV grid resolution. Intuitively, one pixel in the feature map of $k$-th level corresponds to one grid of depth $z_k$.
% % 
% \begin{table}
% \setlength{\tabcolsep}{8pt}
% \begin{center}
% \label{table:multi-scale}
% \vspace*{1mm}
% \begin{tabular}{ccccc}
% \hline
% Notation & $\bf F_1$ & $\bf F_2$ & $\bf F_3$ & $\bf F_4$ \\ 
% \hline
% Feature Scales & 1/8 & 1/16 & 1/32 & 1/64  \\
% \hline
% Depth Intervals & (39m, 50m] & (19m, 39m] & (9m, 19m] & [1m, 9m] \\
% \hline
% \end{tabular}
% \vspace*{-6mm}
% \caption{Depth intervals assignment for different scales.}
% \end{center}
% \end{table}

% \subsection{Two-stage transformation from Perspective to BEV View}
\subsection{Geometry-guided Pre-Alignment}
\label{text:GPA}
% /Geometry-guided Consistency Learning}
\begin{figure}[t]
	\begin{center}
		\includegraphics[width=1\linewidth]{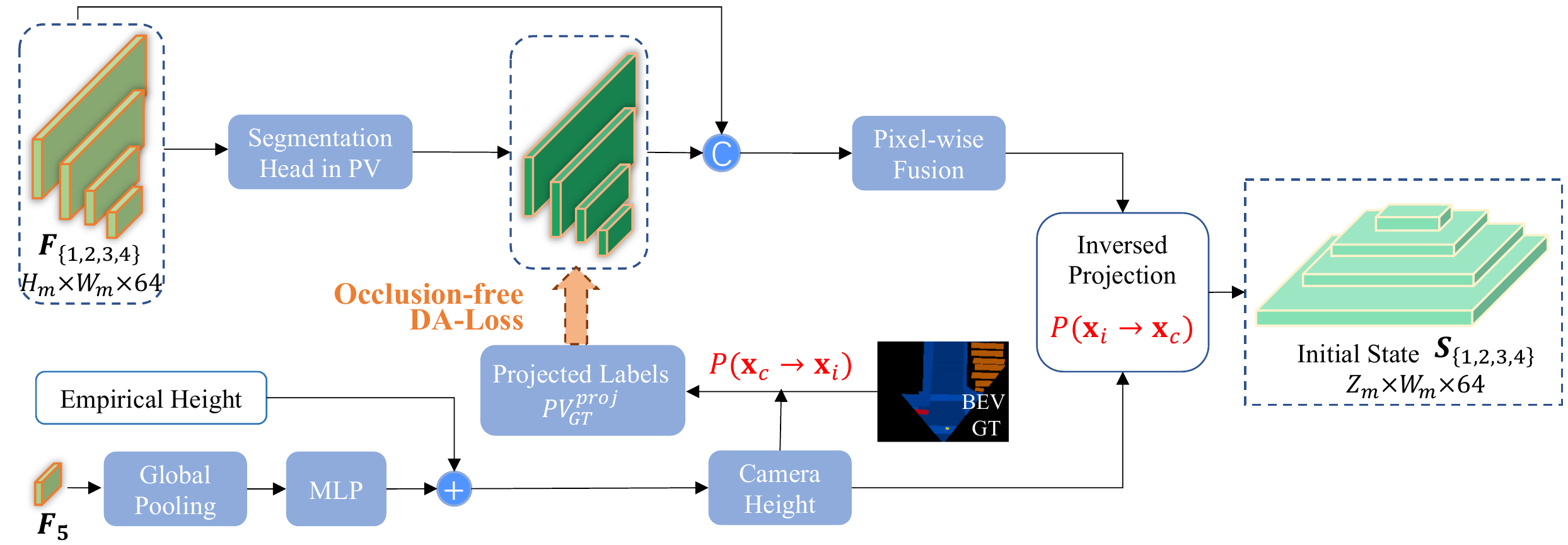}
	\end{center}
	%\vspace*{-5mm}
	\caption{Geometry-guided Pre-Alignment Module. The pyramid image features are first fed into the segmentation head to predict the \Xiaoqing{BEV-consistent} probability maps \Xiaoqing{enforced by the} \textit{Occlusion-free DA loss} with \textit{projected labels}. \Xiaoqing{The BEV-consistent probability maps and the perspective features are further encoded by pixel-wise fusion to extract visibility-aware features.}
	In the other branch, \Xiaoqing{the smallest-scale} features {$\boldsymbol{F}_5$} are used to predict the offset \Xiaoqing{w.r.t. to the empirical predefined height}. Then the learned camera height is applied to inversely project the visibility-aware perspective features to BEV features, which serve as initial queries of the follow-up transformer stage. 
	}
	%}
	\label{fig:step1}
\end{figure}

In this section, we introduce \Xiaoqing{the first stage, \textit{i.e.}, the Geometry-guided Pre-Alignment module. We first present}
the geometric relation between the perspective view and BEV. Then we detail the consistency between image features and projected BEV labels, and describe our visibility-aware feature learning method. Finally, we describe geometry-based warping to obtain the pre-aligned BEV features. \Xiaoqing{The detailed pipeline of this module is depicted in Figure \ref{fig:step1}.}

\noindent\textbf{Learnable Geometric Relation.}
The transformation from perspective view (PV) to BEV can be given by a \Xiaoqing{projection} matrix $P$.
\Xiaoqing{We first introduce the coordinate systems: a certain point in the camera coordinate system is represented by $\bold{x}_c=[x_c, y_c, z_c]^T \in \mathbb{R}^3$}. The ground space is by setting the y-coordinate of the camera coordinate system to $h$ and a certain point lying on the ground plane turns out to be $\bold{x}_c=[x_c, h, z_c]^T$, where $h$ denotes the height of the mounted camera from the ground. The BEV coordinates simply remove the $y$-dim and can be denoted as $\bold{x}^B=[x_c,  z_c]^T \in \mathbb{R}^2$. In the following, we do not particularly distinguish the BEV space from the ground space in the camera coordinate system.
The homogeneous image coordinates $\bold{x}_i=[x_i, y_i, 1]^T$ have a one-to-one correspondence with the ground coordinates, which can be expressed by:
\begin{equation}
\label{eq:P}
    P(\bold{x}_c \xrightarrow{} \bold{x}_i): \quad\bold{x}_i = K\bold{x}_c/z_c=K[x_c/z_c, h/z_c, 1]^T
\end{equation}
where $K$ is the camera intrinsic matrix: $K=[[f_x, 0, c_x], [0, f_y, c_y], [0, 0, 1]]^T$, and the inverse transformation from image to ground coordinates is formulated as:
% \begin{equation}
% \label{eq:P_i}
% \begin{split}
%  P(x_i\xrightarrow{} x_c):\quad x_c =\frac{(x_i-c_x)z_c}{f_x} ,\\
%  P(y_i\xrightarrow{} z_c):\quad z_c =\frac{f_y h}{y_i-c_y}
%   \end{split}
% \end{equation}

\begin{equation}
\label{eq:P_i}
P(\mathbf{x}_i \to \mathbf{x}_c):
\left\{\begin{matrix}
 \quad x_c =\frac{(x_i-c_x)z_c}{f_x}\\
z_c =\frac{f_y h}{y_i-c_y}
\end{matrix}\right.
\end{equation}

Based on the geometric correspondences illustrated  in the Equation (\ref{eq:P}) and (\ref{eq:P_i}), we \Xiaoqing{are able to transform from the perspective space to the BEV.}
% , refer to the red-colored region in Figure \ref{fig:step1}.}
% 
In this way, we are able to recover the coarse ground coordinates given the image coordinates and the camera height. However, as is acknowledged in \cite{roddick20_PON}, the camera height $h$ is unavailable for a real monocular perception system.
% , increasing the challenge of the task. 
Alternatively, we enforce the network to learn the camera height parameters.
% given a single image. 
The image features ${\boldsymbol {F_5}}$ with the scale of $\times 1/128$ are compressed into a vector by global average pooling and followed by an MLP to leverage the global context for predicting the offset of the camera height to the empirically predefined height.
% 
% \old{\noindent\textbf{Consistency between Image Features and Projected labels.}}

\Xiaoqing{\noindent\textbf{Visibility-aware Perspective Feature Learning.} The BEV semantic segmentation is an implicit mapping-segmentation coupling task. 
% Due to the gap between different spaces, end-to-end features learning for BEV segmentation only under the supervision of BEV ground truth (GT) is non-trivial.
Here we decouple the BEV segmentation into the geometric prior-based mapping and perspective segmentation.
The latter is supervised by an explicit segmentation loss with projecting the BEV GT labels onto the image plane following the transformation $P({\boldsymbol{ x_c}}\xrightarrow{} {\boldsymbol {x_i}})$ in the Equation (\ref{eq:P}) to generate the projected labels ${PV}^{proj}_{gt}$.}
% \old{The \textit{projected labels }are produced by projecting the BEV semantic labels onto the image plane following the transformation $P({\bold x_c}\xrightarrow{} {\bold x_i})$ in the Equation (\ref{eq:P}).} 
${PV}^{proj}_{gt}$ reflects the whole perspective-view ground including visible or invisible regions. However, the perspective features extracted from images only reflect the visible  foreground regions.
Therefore, the projected labels ${PV}^{proj}_{gt}$ can be used to obtain the visibility-aware image features by fusing the information of projected labels and image features.
In specific, the pyramid features $\boldsymbol {F_{\{1,2,3,4\}}}\in {\mathbb{R}^{H_i\times W_i\times 64}}$ are separately fed into the weight-shared segmentation head to generate the \Xiaoqing{corresponding} probability maps $\boldsymbol {P_{\{1,2,3,4\}}}\in {\mathbb{R}^{H_i\times W_i \times C}}$ under the supervision of our depth-aware Dice (DA-Dice) segmentation loss with projected labels.
% \Xiaoqing{In this way, the module is supposed to learn BEV-consistent perspective features.} 
% We use an MLP network to modulate the image features conditioned on the predicted probability maps by:
We concatenate the feature maps and the corresponding probability maps, and learn the visibility-aware features ${\boldsymbol {A_i}}$ with pixel-wise fusion (MLP) by:
% The inconsistent regions need to integrate more \textit{contextual information} to infer their classes in the BEV. 
% Therefore, we propose the next module to inform the network which image features are inconsistent with projected labels.
% \noindent\textbf{Occlusion-free Projection Supervision.}
% 
\begin{equation}
        {\boldsymbol {A_m}} = {\rm MLP}({\boldsymbol {F_m}}, {\boldsymbol {P_m}})
\end{equation}
\noindent\textbf{Geometry-based Warping.}
From the Equation (\ref{eq:P_i}), we can derive that $z_c / x_c = f_x / (x_i-c_x)$, which indicates that pixels \Xiaoqing{of perspective view} lying on \textit{the same column} (\textit{i.e.}, with the same x-coordinate $x_i$) map onto \textit{the same polar ray} in BEV space with a slope of $f_x / (x_i-c_x)$. 
Following the transformation $P(y_i \xrightarrow{} z_c)$ in  Equation (\ref{eq:P_i}), 
the $\boldsymbol{j}$-th column of augmented image features $\boldsymbol {A_m^j}$ are warped into the BEV space with the learned camera height $h$ \Xiaoqing{by inverse projection}. That is:
\begin{equation}
\begin{split}
    {\boldsymbol {S_m^j}} = {\rm Warp}( {\boldsymbol  {A_m^j}}; P(y_i\xrightarrow{} z_c))
\end{split}
\end{equation}
where $\{\boldsymbol{S_m^1, S_m^2,...,S_{m}^{W_m}}\}$  are computed in parallel in our implementation, and are concatenated to output the tensor ${\boldsymbol {S_m}}\in \mathbb{R}^{Z_m \times W_m\times 64}$. 
\Xiaoqing{The warped BEV features take advantage of both the appearance from multi-scale perspective-view features and the visibility with BEV projected-to-PV labels guidance. The geometry-guided transformation module provides initial queries for the follow-up transformer stage for further tuning the features for the BEV segmentation task.}

% \begin{figure}
% \centering
% \includegraphics[height=3.2cm]{figs/CPB.pdf}
% \caption{Ground prior-based transformation Module. The pyramid image features are first fed into the segmentation head to predicting the probability map under the  supervision of projected semantic labels in BEV space. \Xiaoqing{TODO}}
% \label{fig:gpb}
% \end{figure}
\subsection{Ray-based Transformer}
% /Consistency Informed Transformer}

\begin{figure}[t]
	\begin{center}
		\includegraphics[width=1\linewidth]{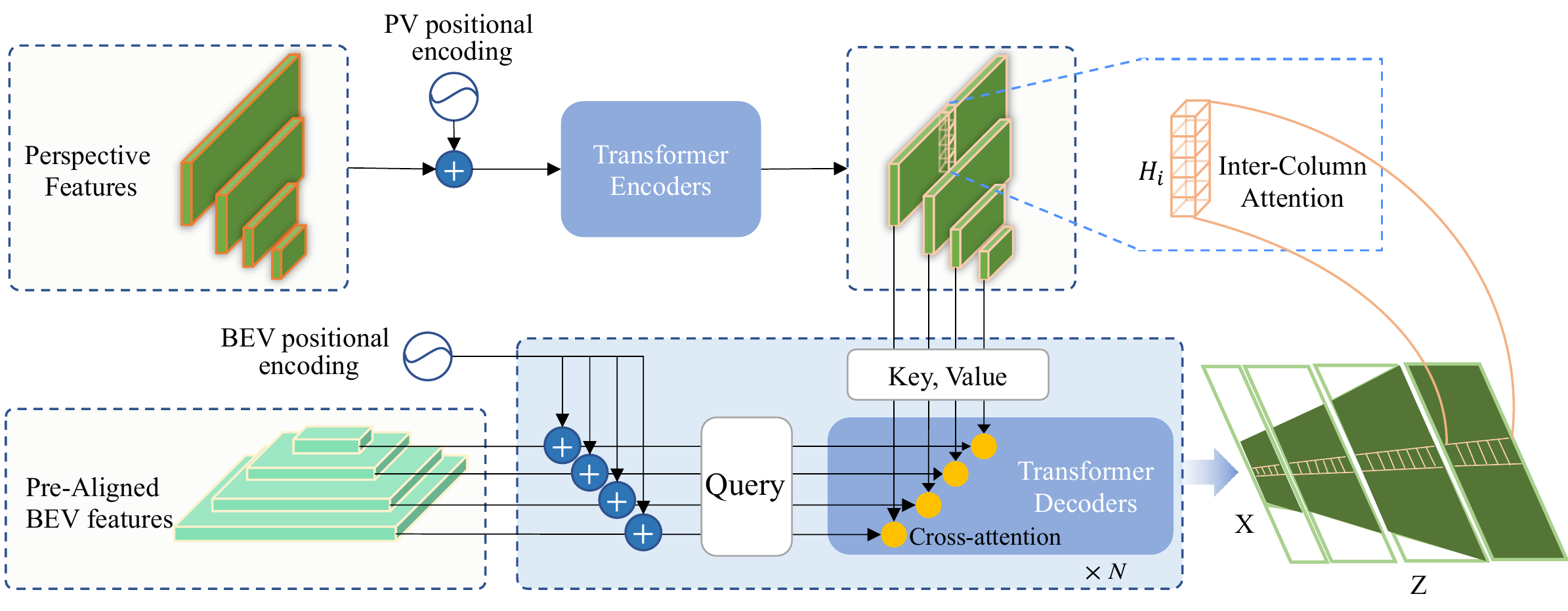}
	\end{center}
	%\vspace*{-5mm}
	\caption{Ray-based Transformer. The \Xiaoqing{pyramid} perspective features, along with the positional encoding in perspective view, are fed into the transformer encoders \Xiaoqing{to integrate the knowledge within the same column.} In the next decoder module \Xiaoqing{where the inter-column cross-attention is conducted}, the output features of encoder serve as \textit{Key} and \textit{Value}, and the pre-aligned BEV features, along with the BEV positional encoding, work as \textit{Query}. The initial queries  ${\boldsymbol{\widetilde{S}_m}}$ are refined by $N\times$ \Xiaoqing{stacked} decoder layers. Finally, the refined BEV features ${\boldsymbol{\widetilde{S}_m}} \in \mathbb{R}^{ Z_m\times W_m\times 64}$ are warped to $\boldsymbol{M}_m\in \mathbb{R}^{Z_m\times X \times 64}$ in the $X$-$Z$ coordinate system, all of which are concatenated along with ${Z}$ axis to output the BEV feature map $\boldsymbol{M}.$}
	%}
	\label{fig:Step2}
\end{figure}

\Xiaoqing{The second step of our two-stage transformation pipeline is the ray-based transformer, which is depicted in the Fig. \ref{fig:Step2}.}
In this stage, we extend the common multi-head attention \cite{vaswani2017_attention} into our Ray-based Transformer (RT). 
The multi-head attention  needs three inputs of queries $(\boldsymbol{Q})$, keys $(\boldsymbol{K})$, and values $(\boldsymbol{V})$, which is denoted as ${\rm{MultiHead}}(\boldsymbol{Q},\boldsymbol{K},\boldsymbol{V})$. We refer the reader to the literature \cite{vaswani2017_attention} and see the appendix for more detailed descriptions. 
Since our BEV semantic segmentation task requires high-resolution feature maps, computing the attention of a full image, like most existing works, will bring high computation complexity and GPU memory.
% and occupy a huge number of 
% 
As derived in Sec. \ref{text:GPA}, pixels of perspective view lying on the \textit{same column} correspond to the \textit{same polar ray} of birds-eye-view.
This motivates us to compute attention in \textit{a single column or ray}, which greatly reduces the complexity of attention. 

{\noindent \textbf{Differences with the Original Transformer. }} 
Our method draws on the core idea of Transformer, \textit{i.e.}, employing the multi-head attention mechanism. But we have two new designs for our BEV segmentation task. Firstly, We use column-wise attention so that we can perform on high-resolution feature maps, which is indispensable in our pixel-wise recognition. Secondly, we introduce the pre-aligned features encoding the appearance and visibility, along with BEV positional encoding, as queries in the cross attention. 
% \Xiaoqing{We believe the pre-aligned features generated from the geometry-guided pre-alignment module are mostly BEV-consistent. However, the transformer is designed to aggregate image contextual information to further tune the features if inconsistent with the BEV target.} 
The ablation study in the Sec. \ref{text:ablation} validates the superiority of our designs.
In the following part, we detail two attention mechanisms in our transformer, and omit other components such as normalization for the sake of simplicity. The complete structure can be seen in our supplementary material. 

%\textbf{Ray/Column based efficient transformer.}
\noindent\textbf{Column Context Augment(CCA) in Encoder.} 
As illustrated in Fig.\ref{fig:Step2}, the inputs \Xiaoqing{for the transformer encoders} are \Xiaoqing{perspective features} ${\{ \boldsymbol F_1, \boldsymbol F_2, \boldsymbol F_3, \boldsymbol F_4\}}$ extracted from the FPN,
where $\boldsymbol{F_m}$ has a spatial resolution of ${ H_m}\times {W_m}$.
In the CCA, each pixel adaptively integrates the information from other pixels of the same column, by using multi-head self-attention. 
%The attention mechanism is unable to distinguish the position of the different input features. 
% 
We further introduce spatial positional encodings $\boldsymbol{P_m}$ to the input $\boldsymbol{F_m}$ to distinguish the positions of the input features. We use a sine function to generate spatial positional encoding. Let $\boldsymbol {F_m^j}$, $\boldsymbol{P_m^j}$ $\in \mathbb{R}^{H_m \times 64}$ denote the $\boldsymbol {j}$-th column  of $\boldsymbol {F_m}$ and $\boldsymbol {P_m}$, respectively. The mechanism of CCA can be summarized as

\begin{equation}
\begin{split}
\label{equation:Multi-Head Attention}
{\boldsymbol {\widetilde{F}_m^j}} = {\boldsymbol{F_m^j}} + {\rm MultiHead}({\boldsymbol {F_m^j} + \boldsymbol{{P}_{m}^j}},{\boldsymbol {F_m^j} + \boldsymbol{P_m^j}}, {\boldsymbol {F_m^j}}),\\
    {\boldsymbol {\widetilde{F}_m}}={\rm CCA}({\boldsymbol {F_m}})={\rm Concat}({\boldsymbol {\widetilde{F}_m^1}, \boldsymbol {\widetilde{F}_m^2}, ..., \boldsymbol {\widetilde{F}_m^{W_m}}})
\end{split}
\end{equation}

%%%%%%%%%%%%
\noindent\textbf{Ray-based Cross-Attention (RCA) in Decoder.} 
RCA in the transformer decoder aims to refine the output of pre-alignment block,
${{\{\boldsymbol S_1, \boldsymbol S_2,\boldsymbol S_3,\boldsymbol S_4\}}}$, based on the augmented image features ${\{\boldsymbol{\widetilde{F}_1}, \boldsymbol{\widetilde{F}_2}, \boldsymbol{\widetilde{F}_3}, \boldsymbol{\widetilde{F}_4}\}}$.
As depicted in Fig.\ref{fig:Step2}, the RCA receives the pre-aligned BEV feature as \textit{Query}, the augmented features built from the encoder as \textit{Key} and \textit{Value}. 
% refines the occluded regions, \Xiaoqing{fix this part, about the consistency @gongshi TODO} based on encoding \textit{appearance} and \textit{consistency} information. 
Similar to CCA, spatial positional encoding $\boldsymbol {P_m^{\prime}}$ is also adopted in RCA.  The difference is that $\boldsymbol {P_m^{\prime}}$ represents the position in the BEV, while $\boldsymbol {P_m}$ is on the image plane.
The mechanism of RCA can be summarized as 
\begin{equation}
\begin{split}
% {{\bf{X}}_{cca}} = {\widetilde{\bf{X}}_{cca}} + {\rm{FFN}}\left( {{{\widetilde{\bf{X}}}_{cca}}} \right), \\
{\boldsymbol {\widetilde{S}_m^j}} = {\boldsymbol {S_m^j}} + {\rm{MultiHead}}\left( {\boldsymbol{S_m^j} + {\boldsymbol{{P^{\prime}}_m^j}},{\boldsymbol {\widetilde{F}_m^j}} + {\boldsymbol {P_m^j}},{\boldsymbol{ \widetilde{F}_m^j}}} \right), \\
{\boldsymbol{\widetilde{S}_m}} = {{\rm RCA}({\boldsymbol{ S_m}},\boldsymbol {F_m})} = {\rm Concat}({\boldsymbol {\boldsymbol {\widetilde{S}_m^1}}, \boldsymbol{\widetilde{S}_m^2}, ..., \boldsymbol{\widetilde{S}_m^{W_m}}})
\end{split}
\label{eq-cca}
\end{equation}
% 
% 
% \textbf{Different depth intervals combined to the final map.}
% 
Since the columns of ${\boldsymbol {\widetilde{S}_m}}$ are still in the image coordinate, we warp them to rays in the camera coordinate following the transformation $P(x_i\xrightarrow{}x_c)$ in the Equation (\ref{eq:P_i}) to obtain $\{{\boldsymbol M_1, \boldsymbol M_2, \boldsymbol M_3, \boldsymbol M_4}\}$, which are responsible for different depth ranges. We concatenate all features along the depth axis to obtain the feature maps of the whole scene:
\begin{equation}
\begin{split}
        {\boldsymbol M_m} = {\rm Warp}({\boldsymbol {\widetilde{S}_m}}; P(x_i\xrightarrow{}x_c)),\\
        {\boldsymbol M} = {\rm Concat}({\boldsymbol M_1,  \boldsymbol M_2, \boldsymbol M_3, \boldsymbol M_4})
\end{split}
\end{equation}
The final BEV feature maps $\boldsymbol M$ are fed to the downstream convolutional segmentation network.
\Xiaoqing{Thanks to the CCA and RCA-based transformer, the network take appearance and visibility knowledge into account, and further tunes the invisible regions of the pre-aligned BEV features, based on the context information from the perspective features.
}

% \textbf{Multi-source Cross-attention (borrow more context knowledge rather than naive positional encoding. Pre-alignment helps the transformer to focus  on more detailed prediction. )}

% \textbf{Ray/Column based efficient transformer.}

% 

% \label{text:overview}
% \begin{figure}
% \centering
% \includegraphics[width=1\linewidth]{figs/overview_v3.pdf}
% \caption{Overview}
% \label{fig:overview}
% \end{figure}
%%%%%%%%%%%%%%%%

% \begin{figure}
% \centering
% \includegraphics[height=3.2cm]{figs/RB.pdf}
% \caption{Refinement block}
% \label{fig:rb}
% \end{figure}

\subsection{Loss functions}
\label{text:loss}
The Dice loss is commonly adopted in segmentation for alleviating the data imbalance problem.
%\noindent\textbf{Review of Dice Loss:} 
The GT semantic label of $i$-th pixel in the BEV map is  [$y_i^1$, $y_i^2$,..., $y_i^C$], and the predicted probability is [$p_i^1$, $p_i^2$,..., $p_i^C$], where $y_i^k \in \{0, 1\}$ and $p_i^k \in [0, 1]$, and $C$ is the number of classes.
The dice loss can be formulated as:

\begin{equation}
%  \vspace{0.5cm}
    L_{\rm dice} = 1 - \frac{1}{C}\sum_{k=1}^C{\frac{2\sum_i^N y^k_i p^k_i}{\sum_i^N{y_i^k + p_i^k +  \epsilon}}}
    % \vspace{-0.3cm}
\end{equation}
where $N$ is the number of pixels in a mini-batch and $\epsilon$ is a constant used to prevent division by zero.   

For the BEV semantic segmentation task which is actually an implicit multi-task problem involving 3D location and segmentation, there are two problems that can affect the performance. 
For one thing, due to the perspective projection from the real world to the image plane, distant objects appear to be smaller than nearer objects. In other words, the closer instances occupy much more pixels than farther ones, which dominate the overall segmentation loss in the perspective view.
For another, those pixels that have easily-classified appearances or follow a simple perspective-to-BEV mapping, such as most road regions, comprise the majority of the loss.
% road layouts are easier to be correctly segmented since they are on the ground plane whereas some other objects are hard to be distinguished due to erroneous predicted location or category. 
% \begin{figure}
% \centering
% \includegraphics[height=3.5cm]{figs/DA.pdf}
% \caption{}
% \label{fig:rb}
% \end{figure}

\noindent\textbf{Occlusion-free Depth-aware Dice Loss:}
% \Xiaoqing{@gongshi TODO : copy some descriptions from the former section: projection xx to perspective view.}
As discussed in Sec. \ref{text:GPA}, we project the BEV  labels onto the image plane to generate the projected labels ${PV}^{proj}_{gt}$, which supervises the segmentation on the perspective view.
To tackle the first problem caused by the domination of nearer objects in perspective images, we propose the novel Depth-aware Dice loss by re-weighting the loss in a depth-aware manner. 
% \old{assigning different weights to the loss according to the depth.} 
In specific, the Jacobian determinant gives the ratio of the area ratio between \textit{image ground plane} ($\Delta A_i$) and the \textit{BEV ground plane} ($\Delta A_c$) as:
\begin{equation}
\label{eq:da}
    R_{A_c\xrightarrow{}A_i}=\frac{\partial A_i}{\partial A_c}=|J| = \left|             
  \begin{array}{cc}   
    \frac{\partial x_i}{\partial z_c} &  \frac{\partial x_i}{\partial x_c} \\ [1ex]
   \frac{\partial y_i}{\partial z_c}  &  \frac{\partial y_i}{\partial x_c}  
  \end{array}\right|= \left|                 
  \begin{array}{cc}   
     \frac{-f_x x_c}{z_c^2} & \frac{f_x}{z_c}\\ [1ex]
    \frac{-f_y h}{z_c^2} & 0 
  \end{array}\right|= \frac{f_x f_y h}{z_c^3}
\end{equation}
We find that the area ratio is proportional to $(1/z_c)^3$, thus we re-weight the pixels with the weight $z_c^3$ to solve the imbalance.
The depth-aware dice loss is:

\begin{equation}
    L_{\rm DA\_{dice}} = 1 - \frac{1}{C}\sum_{k=1}^C{\frac{2\sum_i^N z_{ci}^3 y^k_i p^k_i}{\sum_i^N{z_{ci}^3 (y_i^k + p_i^k )+  \epsilon}}}
\end{equation}
\noindent{\textbf{Self-Weighted Dice Loss:} }
To further alleviate the second problem, \textit{i.e.}, the dominating influence from easy samples in training, we propose to associate training samples with dynamically adjusted weights to emphasize hard examples. We first propose a weighting function $I_i^k$ to adjust the hard-mining strength by a parameter $\alpha$ in Equation (\ref{eq:Ik}) and then utilize $I_i^k$ to reweight the Dice loss and obtain the self-weighted dice loss in Equation (\ref{eq:sw_dice}).
\begin{equation}
    I_i^k = 1 + \alpha [y_i^k(1-p_i^k)+(1-y_i^k)p_i^k]_{\rm stop\_grad}
\label{eq:Ik}
\end{equation}
%{We utilize $I_i^k$ to reweight the Dice loss and obtain the self-weighted dice loss as follows: }
\begin{equation}
    L_{\rm sw\_{dice}} = 1 - \frac{1}{C}\sum_{k=1}^C{\frac{2\sum_i^N I^k_i y^k_i p^k_i}{\sum_i^N{I^k_i (y_i^k + p_i^k )+  \epsilon}}}
    \label{eq:sw_dice}
\end{equation}

Note that we detach the weighting function $I_i^k$ to stop the backward propagation of the gradient in Equation (\ref{eq:sw_dice}). Otherwise, the term $p_i^k (1-p_i^k)$ within $I^k_i y^k_i p^k_i$ will be maximized
\Xiaoqing{to make $p^k_i$ fall around the undesired value 0.5.}

%% file: sections/04_Experiments.tex
\section{Experiments}
\label{sec:experiments}

\subsection{Experimental Setup}
\noindent\textbf{Dataset} 
We conduct extensive experiments on two large-scale datasets: The nuScenes \cite{caesar2020_nus} and Argoverse \cite{chang2019_argo} road-scene datasets. 
% 
% The nuScenes dataset consists of 1000 short video sequences collected in Boston and Singapore, using a range of sensors including six surrounding cameras. \Xiaoqing{The dataset includes abundant road semantic  annotations as well as 3D bounding box annotations. }The Argoverse 3D dataset consists of 65 training and 24 validation clips captured in Miami and Pittsburg. 
% 
\Xiaoqing{Since the two datasets are predominantly collected for 3D object detection task rather than BEV semantic segmentation task,} we follow the data generation method in \cite{roddick20_PON} \Xiaoqing{to convert the ground truth 3D bounding box annotations and the vectorized road maps into GT semantic maps in BEV. In addition, for fair comparisons, we also follow the same training and validation splits with other methods.} The nuScenes includes 4 road layout categories and 7 object categories, and the Argoverse includes 7 object categories as well as drivable road. For both datasets, the ground truth of birds-eye-view expands from 1m to 50m in front of the camera \Xiaoqing{\textit{i.e.}, along the $z$-direction} and 25m to either side \Xiaoqing{(\textit{i.e.}, along the $x$-direction)}.  Due to the greater diversity of nuScenes, we choose this dataset for all ablation studies. 
% Like nuScenes, the Argoverse dataset provides both 3D object and road annotations.

\noindent\textbf{Implementation details} 
For fair comparisons, we adopt a pretrained ResNet-50 with a feature pyramid on top as the backbone.
 We adopt a simplified HRNet \cite{wang2020_hrnet} as the BEV segmentation head. In our implementation, we use a simplified HRNet32 by halving the number of blocks in each stage. We use two encoder layers and four decoder layers in the Ray-based transformer. The hyperparameter $\alpha$ \Xiaoqing{ in Equation \ref{eq:Ik}} for the SW-Dice loss is set as 0.5. 
 We adopt a similar depth-interval assignment strategy with \cite{roddick20_PON}, but we only use the former four scales of the FPN.
 \Xiaoqing{The concatenated BEV feature maps from different depth intervals} are of $98\times 100$ pixels, with each pixel covering 0.5m. \Xiaoqing{We obtain the final output map with a resolution of $196\times 200$ pixels by upsampling, which is consistent with other methods.} The model is trained using four Tesla V100 cards, each with 32G memory.
We optimize the network with Adam policy for gradients accumulated over every 8 iterations and train for 40 epochs.
The initial learning rate is set to 0.0002, with a weight decay of 0.99 and batch size 12. 
%which we decay by 0.99 every epoch, and train for 40 epochs. 

% Since both depth intervals along the $Z$ axis and polar rays along the $X$ axis are separately transformed from the image space, there are some discontinuities in the transformed BEV features. These discontinuities can be relieved by contextual information as existing semantic segmentation networks.
% Because of the low resolution of input BEV features map, and small size of objects (pedestrian, traffic cone, etc.), we adopt a simplified HRNet\cite{wang2020_hrnet} architecture that maintains high-resolution representations. 

\noindent\textbf{Evaluation metric} 
Our evaluation metric is the Intersection over Union (IoU) score, which we compute by  binarizing the output probability maps with the threshold of 0.5. Invisible regions are ignored during evaluation following \cite{roddick20_PON}.

\begin{table}[t]
\caption{Results of IoU (\%) on nuScenes validation set. \Xiaoqing{``Mean''} refers to the average IoU over all classes. ``Crossing'': Pedestrian Crossing, ``C.V.'': Construction Vehicle, ``Motor.'':  Motorcycle, ``Ped.'': Pedestrian, ``Cone'': Traffic Cone.}
\label{table:nuscenes}
\begin{center}
\setlength{\tabcolsep}{2pt}
\resizebox{1\linewidth}{!}{
\begin{tabular}{l|cccc|cccccccccc|c}
% \begin{tabu}{l|lllll|lllllllllll|l}
% \tabucline[1pt]{-}
\specialrule{0.6pt}{0.1pt}{0pt}
%\hline
\multirow{2}{*}{Method} & \multicolumn{4}{c|}{Layout} & \multicolumn{10}{c|}{Object} & \multirow{2}{*}{Mean}\\
%\cline{2-17}
 &	{Drivable} & Crossing &	Walkway &	Carpark  &	Bus &	Bike &	Car &	C.V. &	Motor. &	Trailer &	Truck &	Ped. & Cone &	Barrier  &	 \\
 
% \tabucline[1pt]{-}
\hline
\hline
IPM~\cite{roddick20_PON}   &	40.1  &	-  &	14.0  &	-    &	3.0  &	0.2  &	4.9  &	-  &	0.8  &	-  &	-  &	0.6  &	-  &	-  &	- \\
Depth Unpr.\cite{roddick20_PON}   &	27.1  &	-  &	14.1  &	-   &	6.7  &	1.3  &	11.3  &	-  &	2.8  &	-  &	-  &	2.2  &	-  &	-  &	- \\
%\hline
VED \cite{lu2019_VED}  &	54.7  &	12.0  &	20.7  &	13.5    &	0.0  &	0.0  &	8.8  &	0.0  &	0.0  &	7.4  &	0.2  &	0.0  &	0.0  &	4.0   &	8.7 \\
% \hline
VPN \cite{pan2020_VPN} &	58.0  &	27.3  &	29.4  &	12.3   &	20.0  &	4.4  &	25.5  &	4.9  &	5.6  &	16.6  &	17.4  &	7.1  &	4.6  &	10.8    &	17.5 \\
Sim2real \cite{reiher2020_sim2real} &	60.5  &	27.1  &	19.2  &	18.3    & 6.9 &   3.8 &	7.1  &	0.3  &	4.5  &	3.2  &	4.7   &	1.8  &	4.2  &	12.1    &	12.4 \\
% \hline
OFT \cite{roddick2018_OFT} &	62.4  &	30.9  &	34.5  &	23.5    &	23.2  &	4.6  &	34.7  &	3.7  &	6.6  &	18.2  &	17.3  &	1.2  &	1.1  &	12.9    &	19.6 \\
% \hline
PON \cite{roddick20_PON} &	60.4  &	28.0  &	31.0  &	18.4    &	20.8  &	9.4  &	24.7  &	\textbf{12.3}  &	7.0  &	16.6  &	16.3  &	8.2  &	5.7  &	8.1    &	19.1 \\
% \hline
STA-S \cite{saha21_sta}  &	\textbf{71.1}  &	31.5  &	32.0  &	28.0    &	22.8  &	\textbf{14.6}  &	34.6  &	10.0  &	7.1  &	11.4  &	18.1  &	7.4  &	5.8  &	10.8    &	21.8 \\
% \hline
EPOSH \cite{dwivedi2021_EPOSH} &	61.1  &	33.5  &	37.8  &	25.4    &	31.8  &	6.7  &	37.8  &	2.7  &	10.5  &	14.2  &	20.4  &	5.9  &	7.6  &	13.4    &	22.1 \\
% \hline
% TIM \cite{Saha21_tim}  &	\textbf{72.6}  &	36.3  &	32.4  &	30.5  &	43.0  &	32.5  &	15.1  &	37.4  & \textbf{13.8}  &	8.1  &	15.5  &	24.5  &	8.7  &	7.4  &	15.1  &	17.8  &	25.0 \\
\hline
Ours &	65.1  &	\textbf{41.6}  &	\textbf{42.1}  &	\textbf{31.9}    &	\textbf{35.4}  &	13.8  &	\textbf{43.4}  &	9.7  &	\textbf{15.0}  &	\textbf{22.5}  &	\textbf{25.5}  &	\textbf{14.1}  &	\textbf{11.6}  &	\textbf{18.6}   &	\textbf{27.9} \\

% \tabucline[1pt]{-}
\specialrule{0.8pt}{0.1pt}{0pt}
%\hline
\end{tabular}
}
\vspace{-0.8cm}
\end{center}
\end{table}

\begin{figure}
\centering
\includegraphics[height=3.6cm]{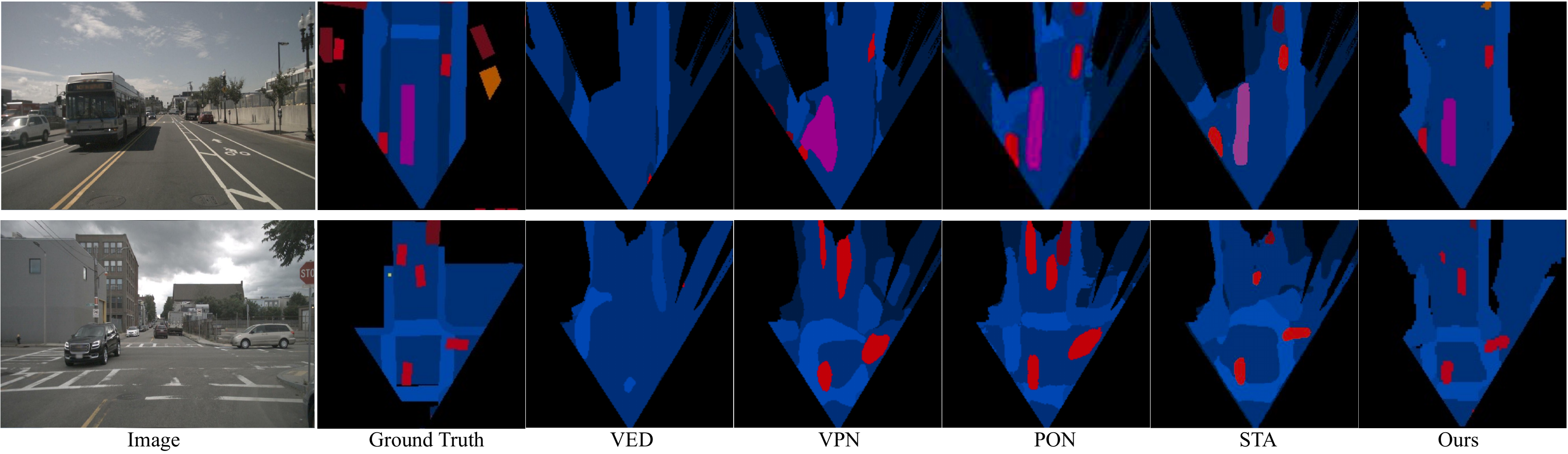}
\caption{Qualitative results on the nuScenes validation set. We compare with the published works and follow their colour scheme.}
\vspace{-0.2cm}
\label{fig:vis}
\end{figure}
\vspace{-0.5cm}

\subsection{Main Results}
We evaluate our method on nuScenes and Argoverse, and compare against the recently published works which belong to different branches: \textbf{(\romannumeral1)} IPM-based methods: IPM~\cite{roddick20_PON}, Sim2real~\cite{reiher2020_sim2real}; \textbf{(\romannumeral2)}
Bottleneck-based methods: VED~\cite{lu2019_VED},  VPN~\cite{pan2020_VPN} and PON~\cite{roddick20_PON};
\textbf{(\romannumeral3)} Depth-based methods: Depth Unprojection-based (Depth-Unpr.)~\cite{roddick20_PON}, OFT~\cite{roddick2018_OFT}, EPOSH~\cite{dwivedi2021_EPOSH} and STA-S~\cite{saha21_sta}. All these works report the results on nuScenes as shown in Table \ref{table:nuscenes}, or provide the results on Argoverse in Table \ref{table:argo}. 
Among all these methods, our method achieves the best performance for most categories and our method surpasses the previous approaches with a significant margin of mean IoU, 6.1\% and 3.2\% on nuScenes and Argoverse, respectively. Fig. \ref{fig:vis} further shows the visual comparisons against other methods on the nuScenes Dataset. 
The two fully-connected bottleneck-based works, VPN and VED, achieve a comparable IoU on the road drivable area,  \Xiaoqing{but  they fail to recognize the smaller objects such as vehicles due to the image features are compressed into a vector. In contrast, our method leverages multi-scale spatial information for different depth intervals  to keep the fine details. For example, as shown in Fig. \ref{fig:vis}, our approach accurately  predicts the vehicles within all depth ranges.}
Compared with other relatively better methods like PON and STA-S, we exploit the geometric prior which helps to accurately locate and identify the road elements, like walkway and pedestrian crossing, which is supported by the qualitative results in Fig. \ref{fig:vis}.

\begin{table}[t]
% \vspace{-0.2cm}
\begin{center}
\setlength{\tabcolsep}{4pt}
\caption{Results of IoU (\%) on the Argoverse validation set.}
\label{table:argo}
\resizebox{1\linewidth}{!}{
% \begin{tabu}{c|c|c|c|c|c|c|c|c|c}

\begin{tabular}{l|c c c c c c c c|c}
% \tabucline[1pt]{-}
\specialrule{0.8pt}{0.1pt}{0pt}
%\hline
Method                  &  Drivable &   Vehicle &   Ped. &   Large veh.    &   Bicycle   &   Bus     &   Trailer &   Motorcy.    &   Mean	    \\    
\hline
\hline
IPM~\cite{roddick20_PON}     & 43.7      &   7.5     &   1.5     &       -       &   0.4         &   7.4     &   -       &   0.8         &   - \\
Depth Unpr.\cite{roddick20_PON}     & 33.0      &   12.7     &   3.3     &       -       &   1.1         &   20.6     &   -       &   1.6         &   - \\
% \hline
VED \cite{lu2019_VED}		&   62.9    &	14.0	&   1.0     &       3.9       &     0.0     &   12.3	&   1.3     &   0.0	        &   11.9         \\
VPN \cite{pan2020_VPN} 	    &   64.9    &	23.9    &	6.2     &	    9.7       &     0.9	    &   3.0	    &   0.4	    &   1.9	        &   13.9        \\
PON \cite{roddick20_PON} 	&   65.4    &	31.4    &	7.4     &	    11.1      &	    3.6	    &   11.0	&   0.7	    &   5.7	        &   17.0        \\
% TIM \cite{Saha21_tim}     &   75.9    &   35.8    &   5.7     &       14.9      &     3.7     &   30.2    &   12.2    &   2.6         &   22.6        \\
\hline
% \hline
Ours	                &   \textbf{67.1} 	&   \textbf{35.9} 	&   \textbf{9.8}    & 	    \textbf{15.7}      & 	\textbf{4.9} 	&   \textbf{31.7} 	&   \textbf{11.3} 	&   \textbf{6.2} 		&   \textbf{20.2}           \\
% \tabucline[1pt]{-}
\specialrule{0.8pt}{0.1pt}{0pt}
% \hline
\end{tabular}
}
\end{center}
\vspace{-0.5cm}
\end{table}

\subsection{Ablation Study}
\label{text:ablation}
We conduct ablation studies to evaluate the key designs in our method. Unless otherwise specified,  we evaluate on the nuScenes validation set. GPA denotes \textit{Geometry-guided Pre-Alignment}, and RT is the \textit{Ray-based Transformer} for short.

\begin{table}[t]
\caption{Effects of different key components. GPA and RT denote the Geometry-guided Pre-Alignment and Ray-based Transformer, respectively. SW and DA refer to the Self-Weight Dice loss and Depth-aware Dice loss.}
\begin{center}
\label{table:ablation}
\setlength{\tabcolsep}{5pt}
\resizebox{0.7\linewidth}{!}{
\begin{tabu}{c|cc|cc|ccc}
% \tabucline[1pt]{-}
\specialrule{0.8pt}{0.1pt}{0pt}
%\hline
\multirow{2}*{Group} & \multicolumn{2}{c|}{Network} &\multicolumn{2}{c|}{Loss}  & \multicolumn{3}{c}{mIoU (\%)} \\
& GPA               &     RT           &   SW         &   DA         & Layout   &  Object  &   Total	\\   
%\hline
\hline
\hline
(a)                  &                     &               &        &   &   31.2    &   4.9  & 12.4 \\
% \tabucline[1pt]{-}
% \hline
 (b)&\checkmark          &                   &               &               & 38.7     &  15.4      &   22.1      \\
 (c)&                    & \checkmark        &               &               & 40.6     &  16.8          &   23.6       \\  
 (d)&\checkmark          & \checkmark        &               &               & 43.2     &  19.1      &   26.0       \\  
% \hline
 (e)&\checkmark          & \checkmark        &\checkmark     &               & 43.8     &  19.9      &   26.7       \\  
 (f)&\checkmark          & \checkmark        &               &\checkmark     & 44.1    &  20.5      &   27.2       \\ 
 (g)&\checkmark          & \checkmark        &\checkmark     &\checkmark     & \textbf{45.2}     &  \textbf{21.0}      &   \textbf{27.9}     \\
% \tabucline[1pt]{-}
\specialrule{0.8pt}{0.1pt}{0pt}
%\hline
\end{tabu}
}
\end{center}
\vspace{-1cm}
\end{table}

\noindent\textbf{Effects of different components.} To analyze the effects of the key designs, we try different combinations and summarize the ablation results in Table \ref{table:ablation}.
\begin{itemize}
\vspace{-2mm}
\item[$\bullet$]\textbf{Baseline.} Group (a) is the baseline that is similar to \cite{reiher2020_sim2real}. We transform the image features onto the ground plane via a homography matrix. The difference between it and our {GPA} is that it adopts a fixed camera height and is not supervised by the projected semantic maps.
The transformed features are further processed by a segmentation network that is the same as our best model for fair comparisons. From Row 1 in Table \ref{table:ablation}, \Xiaoqing{we can see that the baseline achieves reasonable results in road layout areas, but fails to distinguish the objects that standing above the road.}
\item[$\bullet$]\textbf{Network.} 
In Group (b), the {GPA} provides a reliable prior for feature transformation and relieves the effects of occlusion by supervision of projection, improving the mIoU by +9.7\% in total. 
The {RT} (c) transforms the image features to the BEV space by multi-scale column-based attention, which improves the mIoU by +11.2\% in total.
If we combine the {GPA} and {RT} as disscussed in Sec. \ref{text:overview}, their joint effect (d) further enhances the performance by +13.6\% in total. It shows the geometric prior provides complementary information for the {RT}.
\item[$\bullet$]\textbf{Loss.} Groups (e)(f)(g) show the improvements in our proposed loss function. The SW-Dice loss (e) automatically puts higher weights on these pixels that are hard to classify in birds-eye-view, improving the mIoU by +0.7\% in total. The DA-Dice loss (f) balances the pixels of different depth \Xiaoqing{ranges} in the perspective view \Xiaoqing{by reweighting the Dice loss under the guidance of the cubic depth when learning the geometric prior-based pre-alignment module}, which improves the mIoU by +1.2\% in total. The joint of both losses (g) brings a further mIoU gain of +1.9\% in total.
\end{itemize}
\vspace{-0.8cm}
\begin{table}
\begin{center}
\vspace{-0.2cm}
\caption{Effects of components of {GPA}, where ``learnable $h$" denotes learning the jitter of camera height; ``proj. sup." denotes supervising the image features with projected labels \Xiaoqing{from BEV to image space}; ``pixel. fusion" denotes pixel-wise fusion between image features and probability maps of segmentation.}
\label{table:gpb}
\setlength{\tabcolsep}{5pt}
\resizebox{0.85\linewidth}{!}{
\begin{tabu}{c|ccc|ccc}
% \tabucline[1pt]{-}
\specialrule{0.8pt}{0.1pt}{0pt}
%\hline
{Group} & {learnable} {$h$}   &   {proj. sup.}    &   {pixel. fusion}       & {Layout}    &   {Object}  &   {Total}\\
\hline
\hline
\uppercase\expandafter{\romannumeral1} &                &                   &               &   40.2        &    16.7     &   23.4  \\
\uppercase\expandafter{\romannumeral2} &\checkmark      &                   &               &  41.1          &   17.3      &   24.1  \\
\uppercase\expandafter{\romannumeral3} &\checkmark      &   \checkmark      &               &  42.6          &   18.3      &   25.2  \\
\uppercase\expandafter{\romannumeral4} &\checkmark      &   \checkmark      &   \checkmark  &  \textbf{43.2 }         &     \textbf{19.1}      &   \textbf{26.0}  \\
% \tabucline[1pt]{-}
%\hline
\specialrule{0.8pt}{0.1pt}{0pt}
\end{tabu}
}
\end{center}
\vspace{-0.5cm}
\end{table}
\vspace{-0.5cm}
\noindent\textbf{Effects of components of GPA:} Three key designs are presented in Geometry-guided Pre-Alignment to \Xiaoqing{better convert the perspective image features to BEV features, including the learnable camera height, the projection supervision, and pixel-wise fusion between probability maps and image features}.
\Xiaoqing{Comparing Group I and II, where we enforce the network to learn the offset of the camera height, we observe a 0.7\% mIoU gain. 
Group III leverages the projected labels from BEV space to image space to supervise the feature learning procedure, which further improves the performance by 1.1\% mIoU.  The further pixel-wise fusion between perspective features and segmentation probability maps in Group IV further lifts the performance, resulting in a total of 2.6\% gain.}

\begin{table}[t]
    \begin{subtable}[h]{0.46\textwidth}
    % \vspace{-0.2cm}
    \caption{Effects of hyperparameter $\alpha$}
        \centering
        \setlength{\tabcolsep}{3pt}
        \begin{tabu}{c|ccccc}
        \specialrule{0.8pt}{0.1pt}{0pt}
        %\hline
        % \tabucline[1pt]{-}
        $\alpha$    &   0   &   0.25    &   0.5   &   1.0  &    2.0     \\
        \hline
        \hline
        Layout  &       44.1    &      44.2     &      45.2     &    \textbf{45.3}   &     45.1        \\
        Object  &     20.5      &    20.9       &     \textbf{21.0}      &    20.9   &     20.6        \\
        Total  &     27.2      &     27.5      &      \textbf{27.9}     &    \textbf{27.9}   &      27.6       \\
        % \tabucline[1pt]{-}
        %\hline
        \specialrule{0.8pt}{0.1pt}{0pt}
        \end{tabu}
       \label{tab:sw}
    \end{subtable}
    \hfill
    \begin{subtable}[h]{0.46\textwidth}
    \caption{Effects of decoder layers in RT}
        \centering
        \setlength{\tabcolsep}{3pt}
        \begin{tabu}{c|ccccc}
        \specialrule{0.8pt}{0.1pt}{0pt}
        % \tabucline[1pt]{-}
        % \hline
        layers   &  0    & 1   &   2   &   3 &4 \\
        \hline
        \hline
        Layout  &  38.7    &43.7  &   44.1  & 44.8 &  \textbf{45.2}  \\
        Object  &  15.4    &19.2  &   20.8  & \textbf{21.2} &  21.0  \\
        Total   &  22.1    &26.2  &   27.5  & \textbf{27.9} &  \textbf{27.9}  \\
        % \tabucline[1pt]{-}
        \specialrule{0.8pt}{0.1pt}{0pt}
        %\hline
        \end{tabu}
        \label{tab:rft}
     \end{subtable}
     \caption{Effects of $\alpha$ \Xiaoqing{in the proposed SW-Dice loss and the number of } decoder layers \Xiaoqing{within the ray-based transformer module.}}
     \label{tab:temps}
    %   \vspace{-0.5cm}
\end{table}

\noindent\textbf{Effects of $\alpha$ \Xiaoqing{in SW-Dice loss}:} 
%Results of different $\alpha$ in SW Dice loss are shown in Table \ref{tab:sw}. 
The SW-Dice loss introduces the hyperparameter $\alpha$ to control the strength of the modulating term with respect to the predicted probability.
As is shown in Table \ref{tab:sw},  $\alpha=0$ means our loss is equivalent to the plain Dice loss. As $\alpha$ increases, the predicted probability gets dominant in the weighting function. Under all settings of $\alpha$, the proposed SW-Dice loss stably outperforms the baseline ($\alpha=0$). With the best setting, the SW-Dice loss yields a 0.7\% improvement over the plain Dice loss.

\noindent\textbf{Effects of decoder layers in RT:} Table \ref{tab:rft} shows the performance with various number of decoder layers \Xiaoqing{within the ray-based transformer}. Our model can yield 4.1\% improvement even using one layer. The gain reflects that the pre-aligned BEV features can provide a good initialization for the decoder. 
% \Xiaoqing{Why, you should explain in one sentence @gongshi}. 
With the decoder layers increasing, higher performance can be achieved. We observe that it becomes saturated when adopting more than three layers.

\begin{figure}
\centering
\includegraphics[height=3.8cm]{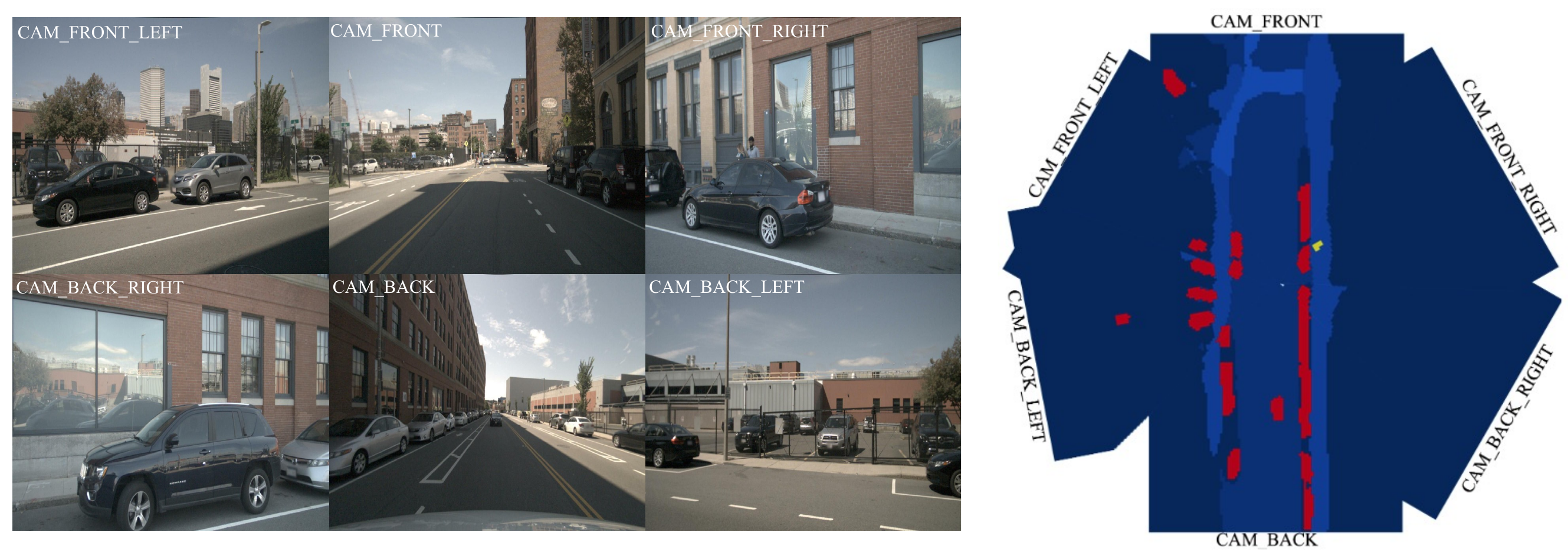}
\caption{An example of late-fusion of six surrounding birds-eye-view semantic maps, which predict consistent full 360$^\circ$ BEV semantic maps.}
\label{fig:stitch}
\vspace{-0.4cm}
\end{figure}
\vspace{-0.4cm}

\subsection{Multiple Views Fusion}
\Xiaoqing{Due to the limited field of view (FOV) of a single camera, it is essential to make full use of all surrounding cameras from multi-view  to perceive the integrated scope of the scene.}
For this purpose, we introduce a late-fusion technique based on Bayesian filtering \cite{roddick20_PON,thrun2002_prob}. Suppose that $R_i \in \mathbb{R}^{2\times 2}$ and $t_i \in \mathbb{R}^{2\times 1}$  are the BEV rotation and translation matrix of $i$-th camera with respect to the ego car coordinates. Let $O_i$  denote the predicted logits (before the sigmoid activation) in $i$-th view.
$O_i$ is warped to the car coordinate system, and we sum over all warped logits maps. The sum of logits are normalized by the sigmoid function $\sigma$ to output the fused probability map $P_{\rm fuse}$.
% \begin{equation}
% \label{eq:fuse}
%     P_{\rm fuse} = \sigma(\sum_{i=1}^V {\rm Warp}(O_i; R_i|t_i))
% \end{equation}
In Fig. \ref{fig:stitch}, we give an example of the fused 360$^\circ$ BEV semantic maps from six surround-view cameras. \Xiaoqing{It validates that our approach  can be applied seamlessly }to predict consistent maps across views.

%% file: sections/05_Conclusion.tex
\section{Conclusion}
In this paper, we proposed a novel method \ourMethod{} for predicting \Xiaoqing{semantic} birds-eye-view maps from monocular images. 
% \old{Our approach improves on the state-of-the-art by incorporating a two-step transformation layer, which includes a \textit{Ground Prior Block}, that provides a coarse BEV features based on the ground prior, and a \textit{Ray-based Feature Transformer}, which updates the coarse features based on column context. As well as predicting BEV maps from a single view, our method can be easily extend to build a full-scene BEV map from all surrounding views.}
% 
\Xiaoqing{The \ourMethod{} leverages a two-stage pipeline to transform the perspective view into the birds-eye-view, which first performs geometry-guided pre-alignment and then further enhances the BEV features based on ray-based transformers. Our approach can also be easily adapted to multi-view scenarios to build a full-scene BEV map.

\textbf{Acknowledgments.} This research was supported by the National Key Research and Development Program of China under Grant No. 2018AAA0100400, the National Natural Science Foundation of China (62176098, 61703049) and the Natural Science Foundation of Hubei Province of China under Grant 2019CFA022.
% \noindent{\textbf{Limitation.}} 
% In order to reduce computation head, adopts the efficient ray-based attention mechanism that we compute the attention map in a single column. In cases of far-away or seriously occluded objects where no context information can be reflected from a certain column in the perspective view, missing parts of objects can happen in the BEV map. In future work, we will introduce the pre-defined categorical tokens to every column so that our ray-based transformer can dig out the information of missing parts.
}

%% file: sections/06_Appendix.tex
\appendix

\section{Ray-based Transformer}
\subsection{Detailed architecture}
\begin{figure}[h!]

	\begin{center}
		\includegraphics[width=0.7\linewidth]{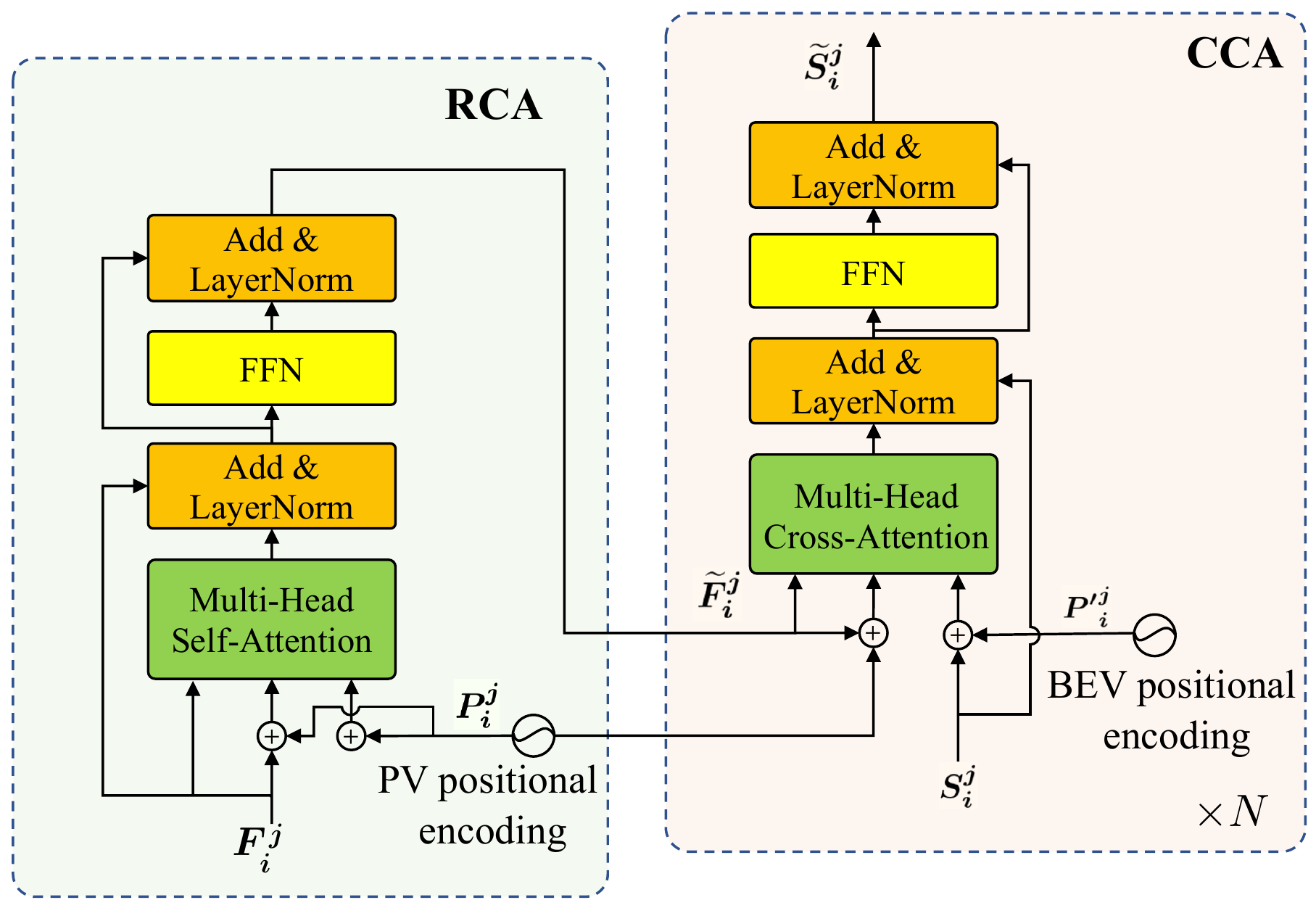}
	\end{center}
	\caption{Architecture of Ray-based Transformer.
	}
	\label{fig:transformer}
\end{figure}

The detailed description of the Ray-based Transformer adopted in GitNet, with positional encodings passed at each attention layer, is given in Fig.~\ref{fig:transformer}.
The perspective image features $\boldsymbol{F_i^j}$, \textit{i.e.}, the $\boldsymbol j$-th column of the $\boldsymbol i$-th level of pyramid features, are passed
through the transformer encoder (Column Context Augment, CCA), together with perspective positional encoding
that are added to queries and keys at every multi-head self-attention layer.
Then, the decoder (Ray-based Cross-Attention, RCA) receives queries, that are initialized as 
the pre-alined features $\boldsymbol{S_i^j}$, along with the BEV positional encoding, and the output of encoder $\boldsymbol{\widetilde{F}_i^j}$, along with the perspective positional encoding, and produces the refined features $\boldsymbol{\widetilde{S}_i^j}$ through multi-head cross-attention.

\subsection{Positional Encoding}
\textbf{2D positional encoding} As the transformer is unable to distinguish the position of elements, we add positional encoding to the \textit{Keys} and \textit{Queries} following \cite{carion2020_detr,vaswani2017_attention}. 
% Suppose that $d$ is the input dimension ($d=64$ in our implementation),
The 2D positional encoding map have the same shape with the input features, that denotes as $P \in \mathbb{R}^{H\times W \times d}$, where $d$ is the channel number, $H$ and $W$ are the height and width of input features, at horizontal and vertical direction, respectively.
We encode horizontal position in the first half of $d$ channels, and the vertical positions in the second half channels.
Suppose the $u\in [0, W)$ and $v\in [H)$ denotes the row and column index, then the horizontal positional encoding at the point of $(u, v) $is:
\begin{equation}
\begin{split}
    P_{\rm h}(u, v, 2i)={\sin}(u/10000^{4i/d})\\
    P_{\rm h}(u, v, 2i+1)={\cos}(u/10000^{4i/d})
\end{split}
\end{equation}
The vertical positional encoding is:
\begin{equation}
\begin{split}
 P_{\rm v}(u, v, 2i+d/2)={\sin}(v/10000^{4i/d})\\
    P_{\rm v}(u, v, 2i+1+d/2)={\cos}(v/10000^{4i/d})
\end{split}
\end{equation}
Where $i\in[0, d)$ is the channel index. We concatenate the positional encoding at two directions to get the whole 2D positional encoding, that is $P={\rm Cat}(P_{\rm h}, P_{\rm v})$.
\noindent\textbf{PV and BEV positional encoding} Both our perspective-view (PV) and birds-eye-view (BEV) positional encoding follow the paradigm of 2D positional encoding detailed above. 
The only difference is that PV positional encoding has the same spatial size with image features, while the BEV positional encoding has spatial size of rasterized BEV map.
% In our ray based transformer, we add perspective positional encoding to the image features, and birds-eye-view positional encoding to the BEV features.

\begin{figure}[h!]

	\begin{center}
		\includegraphics[width=1\linewidth]{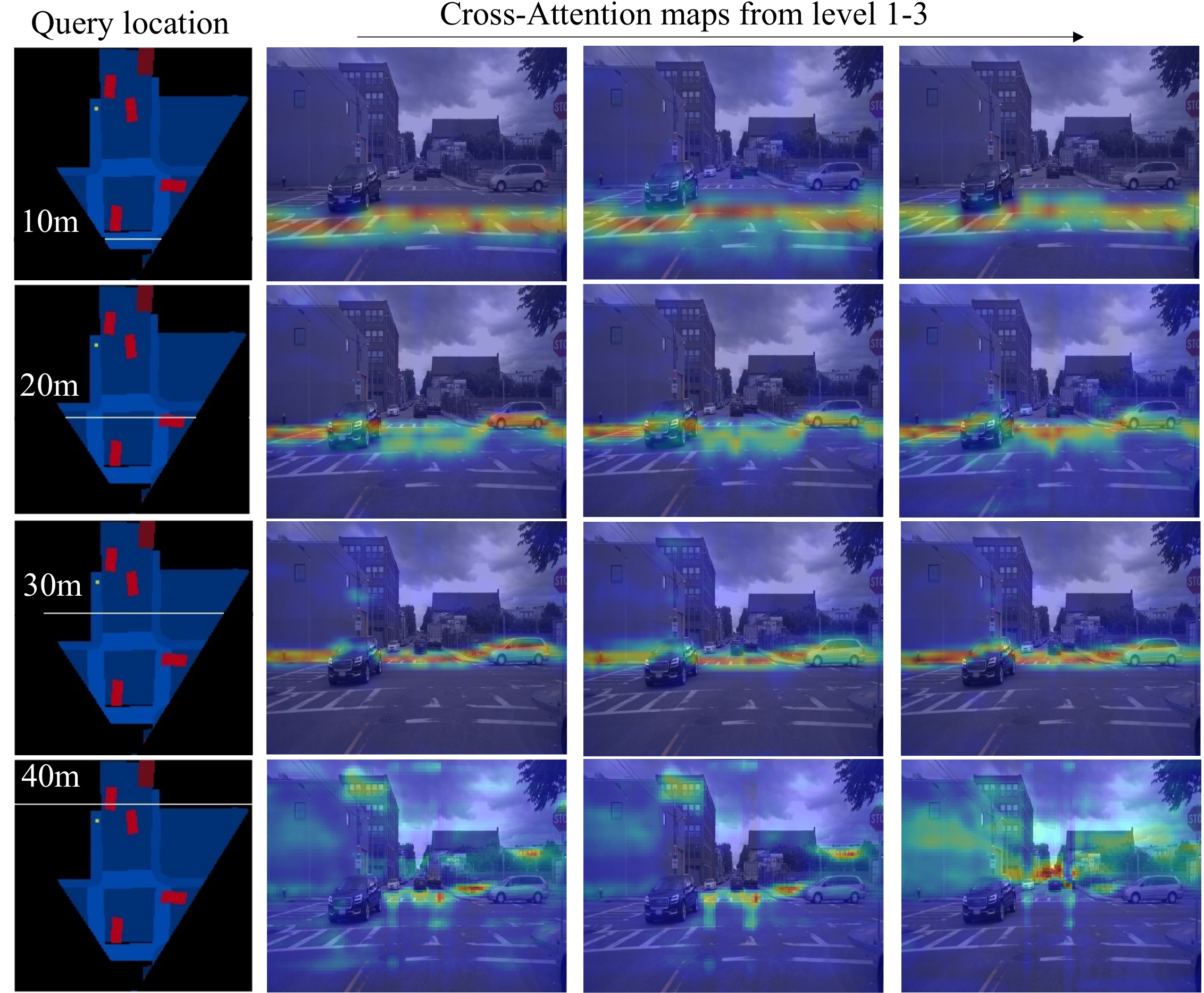}
	\end{center}
	\caption{Visualization of cross-attention map from three levels of decoder layers. In the first column, the white line marked in BEV semantic labels denotes the queries location at 10, 20, 30 and 40 meters away from the camera. The three columns on the right denote
	the cross-attention map from the level-1 to level-3 of the transformer decoder layers. For intuitive comparison, we superimpose the input RGB images onto the cross-attention maps.  
	}
\label{fig:att_map}
\end{figure}

\subsection{What does the cross-attention see?}
To explore how the cross-attention module works in our framework, we visualized the attention maps of a representative sample, as shown in Fig. \ref{fig:att_map}. As our ray-based transformer computes the cross-attention between every query point in the BEV and the corresponding column of the perspective image features, a lateral query line cross all columns/rays (white query line in Fig. \ref{fig:att_map}) produces the attention maps of the full perspective image features.
We depict the cross-attention map from three different decoder layers, which go deeper from left to right. In the first row, the queries lie on the pedestrian crossing, 10 meters away from the camera, and in the right three columns, we can observe that the corresponding cross-attention maps mainly focus on the pedestrian crossing region of the perspective space. When the queries line move farther,20 meters away, the attention maps focus on upper regions of the perspective images. 
Since our pre-alignment module provides visibility-aware pre-aligned features to the transformer, the invisible regions can be further refined by aggregating contextual information from other visible regions. This can be supported by the observation that the attention maps of invisible regions tend to disperse over an extensive region, while attentions maps of visible ground mostly focus a certain point.
In the fourth row, the query line is 40 meters away, and the attention maps are scattered in the invisible regions that occluded by the building and cars.

\begin{figure}[h!]

	\begin{center}
		\includegraphics[width=1\linewidth]{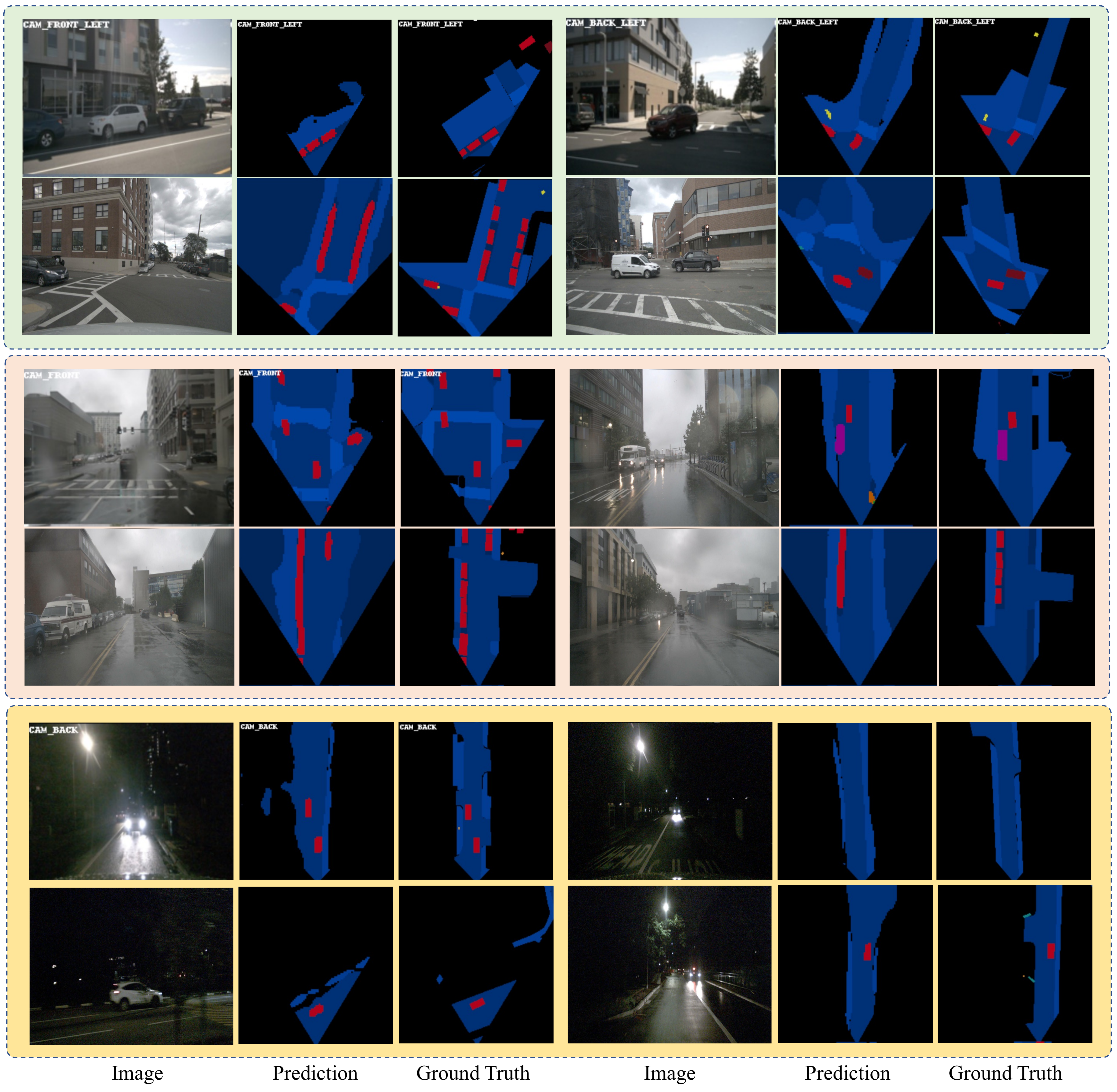}
	\end{center}
	\caption{Visualization of samples under three conditions including light, rainy and dark.}
\label{fig:vis}
\end{figure}

\section{Robustness under Different  Conditions }
In Fig. \ref{fig:vis}, we depict additional visualization results by selecting samples from the validation set of nuScenes including three different weather conditions: light, rainy and dark. For the purpose of practical use, our model must be able to handle these various conditions. From the Fig. \ref{fig:vis}, our model can perform well in the light conditions (the first group), and under the more challenging rainy condition (the second group), our model can segment almost all other cars and the complete layout of the crossroads. Under the dark condition (the third group), our model also succeed to segment the forwarding cars, and right sidewalk which cannot be seen clearly even for human.